\documentclass[conference]{IEEEtran}
\IEEEoverridecommandlockouts
\usepackage{cite}
\usepackage{amsmath,amssymb,amsfonts}
\usepackage{graphicx}
\usepackage{textcomp}
\usepackage{xcolor}

\usepackage{booktabs}
\usepackage{mathabx}
\usepackage[utf8]{inputenc}
\usepackage{algorithm}
\usepackage{algpseudocode}
\usepackage{subcaption}
\usepackage{multirow}
\usepackage[normalem]{ulem}
\useunder{\uline}{\ul}{}
\newcommand{\re}[1]{{\color{black}#1}}

\def\BibTeX{{\rm B\kern-.05em{\sc i\kern-.025em b}\kern-.08em
    T\kern-.1667em\lower.7ex\hbox{E}\kern-.125emX}}
    
\begin{document}

\title{Efficient Multivariate Time Series Forecasting via Calibrated Language Models with Privileged Knowledge Distillation}

\author{\IEEEauthorblockN{Chenxi Liu$^{1}$, Hao Miao$^{2}$, Qianxiong Xu$^{1, *}$\thanks{* Corresponding author.}, Shaowen Zhou$^{1}$, Cheng Long$^{1}$, Yan Zhao$^{2}$, Ziyue Li$^{3}$, Rui Zhao$^{4}$}

\IEEEauthorblockA{
$^{1}$\textit{S-Lab, Nanyang Technological University, Singapore}\\
$^{2}$\textit{Aalborg University, Denmark}
$^{3}$\textit{University of Cologne, Germany}\\
$^{4}$\textit{SenseTime Research, China}\\
{\{chenxi.liu, qianxiong.xu, shaowen.zhou, c.long\}@ntu.edu.sg,} \\ 
{\{haom, yanz\}@cs.aau.dk,} {zlibn@wiso.uni-koeln.de, zhaorui@sensetime.com}
}}

\maketitle

\begin{abstract}
Multivariate time series forecasting (MTSF) endeavors to predict future observations given historical data, playing a crucial role in time series data management systems. With advancements in large language models (LLMs), recent studies employ textual prompt tuning to infuse the knowledge of LLMs into MTSF. However, the deployment of LLMs often suffers from low efficiency during the inference phase. To address this problem, we introduce TimeKD, an efficient MTSF framework that leverages the calibrated language models and privileged knowledge distillation. TimeKD aims to generate high-quality future representations from the proposed cross-modality teacher model and cultivate an effective student model. The cross-modality teacher model adopts calibrated language models (CLMs) with ground truth prompts, motivated by the paradigm of Learning Under Privileged Information (LUPI). In addition, we design a subtractive cross attention (SCA) mechanism to refine these representations. To cultivate an effective student model, we propose an innovative privileged knowledge distillation (PKD) mechanism including correlation and feature distillation. PKD enables the student to replicate the teacher's behavior while minimizing their output discrepancy. Extensive experiments on real data offer insight into the effectiveness, efficiency, and scalability of the proposed TimeKD. 
\end{abstract}


\section{Introduction}
\label{sec:intro}


The rapid revolution of sensing techniques and the proliferation of edge devices generate massive amounts of time series data~\cite{campos2023lightts,liu2024mvcar,miao2024less,wang2023observed}. It is important to extract useful knowledge from the historical time series data enabling a variety of time series-related applications, such as arrival time estimation~\cite{tran2020deeptrans,yuan2020effective,liu2021understanding}, traffic prediction~\cite{DBLP:journals/www/LiuXWCCC22,fang2024efficient,9665313,zhang2025drawing,zhang2024modeling}, and route planning~\cite{xu2024managing,DBLP:journals/www/CaiWCLX24,DBLP:conf/hpcc/ChenWL20}. A fundamental aspect of such applications is multivariate time series forecasting (MTSF), which aims to predict future observations given historical data~\cite{chang2024timedrl,zha2024scaling,shao2024exploring}. MTSF serves as an essential functionality in the time series data management systems~\cite{sakurai2015mining,DBLP:conf/kdd/Xu0WY024,liu2025vehicle,DBLP:journals/tits/XiaoX0HLZ022,liu2024icde,DBLP:conf/dexa/YangSLLMLZ24}.

Due to its significance, substantial research has been devoted to inventing effective MTSF models~\cite{DBLP:conf/kdd/LiangWNJ0SPW24}. Generally, existing MTSF methods can be categorized into classical methods~\cite{wu2023timesnet,Zeng2022AreTE,Yuqietal-2023-PatchTST,DBLP:journals/corr/abs-2310-06625} and large language model (LLM)-based methods~\cite{DBLP:conf/nips/ZhouNW0023, chang2023llm4ts,sun2023test,xue2023promptcast}. Small classical methods often adopt convolutional neural networks (CNNs)~\cite{wu2023timesnet}, fully-connected networks (FC)~\cite{Zeng2022AreTE}, and Transformers~\cite{Yuqietal-2023-PatchTST} to learn spatial and temporal correlations across variables and time steps for time series forecasting~\cite{miao2025spatio}. However, the limited number of learnable parameters and small-scale training datasets lead these classic methods to extract only shallow features. Recent studies apply large language models for time series forecasting inspired by the fact that time series and language data exhibit similar sequential formats, as well as the generic pre-trained knowledge and knowledge transfer capabilities of LLMs~\cite{DBLP:conf/icde/Trummer24,DBLP:conf/icde/0001JLWC24}. These methods often consider LLMs as deep feature extractors to capture deep features from historical time series facilitating effective time series forecasting.

Unfortunately, due to the huge amount of parameters, the deployment of LLMs (e.g., LLaMA-7B~\cite{touvron2023llama}) would \textit{deteriorate the inference efficiency}, posing challenges to real-time time series forecasting. For example, Time-LLM~\cite{jin2023time}, built on LLaMA-7B, uses 96 historical time steps to predict the next 96 on the ETTm1 dataset, corresponding inference time is 0.135 seconds per sample, 13 times longer than a small classic model, iTransformer~\cite{DBLP:journals/corr/abs-2310-06625}. To address this problem, we propose to adopt knowledge distillation (KD)~\cite{gou2021knowledge} to distill the robust representation capability of large-scale LLMs (teacher) to a small-scale classical model (student), so that the learned student model enables both \textbf{deep feature extraction} and \textbf{fast inference}. Nevertheless, it is non-trivial to develop LLM-guided KD (LLM-KD) methods for MTSF due to the following challenges.

\begin{figure}[t]
    \begin{centering}
    \includegraphics[width=0.48\textwidth]{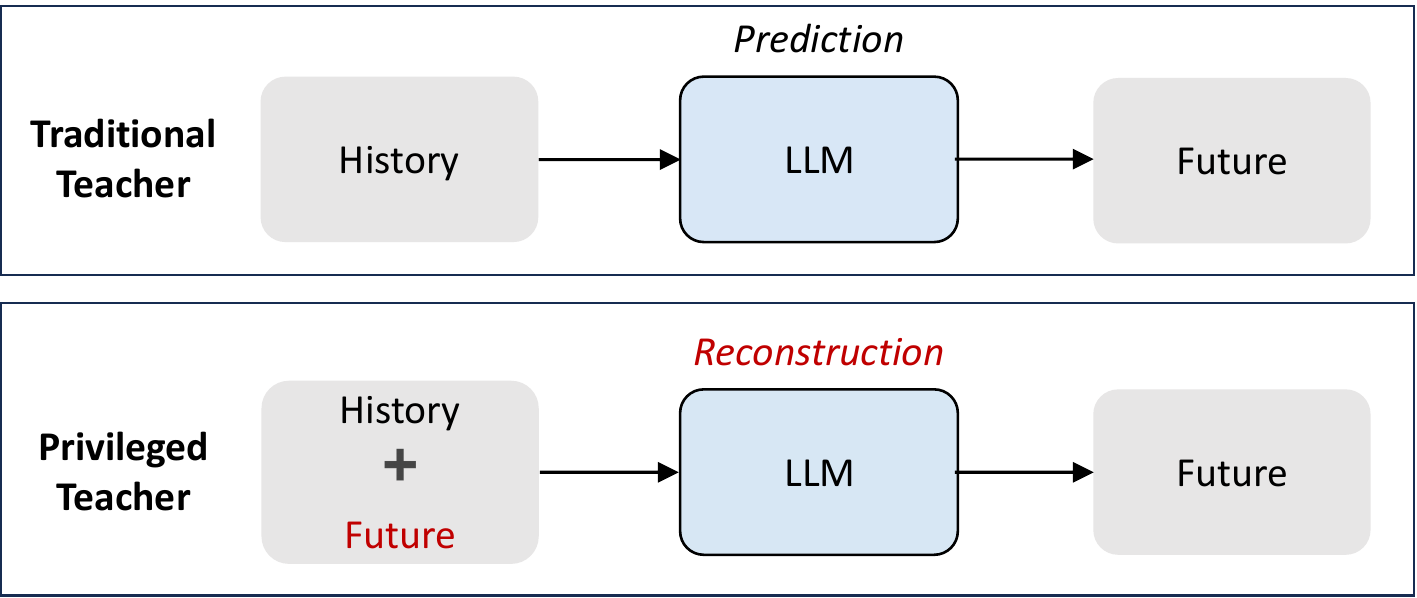}
        \caption{Comparison of traditional teacher models and our privileged teacher. Future data is available only during training and not during testing, thus regarded as privileged information.}
        \label{fig:teachers}
    \end{centering}
\end{figure}

\textit{Challenge I: High-quality Teacher Model Training.} \re{The intuitive approach to designing the teacher model for MTSF tasks is to leverage LLMs to predict future time series, as illustrated in the upper part of Figure~\ref{fig:teachers}.} Specifically, these methods employ traditional teacher models, which take history data as input and produce future features as output~\cite{liu2024taming}. However, this limits the predictive capability of the teacher model due to the constrained distribution of historical data. In addition, the student model may inherit the prediction bias of the teacher model during training. This is particularly evident when dealing with out-of-distribution data~\cite{lu2024diversify}, where the student model's generalization ability may be weaker.

\textit{Challenge II: Effective Student Model Cultivation.} Prevailing LLM-KD methods typically optimize the student model with the output discrepancy of the teacher model. For example, existing LLM-KD methods often rely on black-box distillation~\cite{DBLP:conf/ijcai/LiuHZLM24}, in which the student model only has access to the outputs of the teacher model. These approaches fail to fully capture the teacher model's behavior during the training phase, leading to incomplete knowledge transfer.
Consequently, it is crucial to develop knowledge distillation methods that transfer the training behaviors and outputs from the teacher model to foster effective student model cultivation.

This study addresses the above challenges by providing an efficient multivariate {\ul time} series forecasting framework via calibrated language models with privileged {\ul k}nowledge {\ul d}istillation, entitled TimeKD. To train a high-quality teacher model (\textit{Challenge I}), we develop a cross-modality teacher model for time series reconstruction. Existing methods~\cite{xue2023promptcast,DBLP:journals/corr/abs-2406-01638} demonstrate that wrapping time series as textual prompts enables effective time series feature extraction because the trained LLMs cannot understand pure time series data due to the lack of textual instructions. Building on this insight, We pre-define a prompt template according to the original time series with additional contextual information, e.g., ``\textit{The values were 10, 11, and 20 every hour. Forecast the values for the next 24 hours}''. 

Moreover, we leverage the Learning Under Privileged Information (LUPI)~\cite{DBLP:journals/corr/Lopez-PazBSV15} paradigm in the teacher model. \re{In particular, we treat future data (i.e., ground truth prompts) as privileged information and directly input them into the LLMs to extract effective future representations, as illustrated in the lower part of Figure~\ref{fig:teachers}.} It is notable that taking ground truth as teacher LLM's inputs is feasible for the KD framework without data leakage since the teacher LLMs are only used during training and not during testing. Further, we design calibrated language models (CLMs) to purify the representations of the prompt using a novel calibrated attention mechanism: this mechanism suppresses the inter-modality fusion while enhancing the intra-modality correlations, e.g., the attention scores between text and time series tokens would be biased with a negative value. Subsequently, we design a subtractive cross-attention (SCA)  mechanism to remove the textual information doped in learned future time series representations.

To cultivate an effective student model for time series forecasting (\textit{Challenge II}), we propose an innovative privileged knowledge distillation (PKD). PKD focuses on both correlation and feature distillations to transfer the privileged representations from the LLM-empowered teacher model to a lightweight student model. In terms of correlation distillation, the student model replicates the teacher's behavior based on the shared attention maps of teacher and student models. In terms of feature distillation, the teacher cultivates the student model by minimizing the output discrepancy. By combining these two complementary distillation methods, the teacher model effectively guides the student model in learning internal behavior and enhancing output performance, enabling efficient and effective time series forecasting.

Our primary contributions are summarized as follows.

\begin{itemize}
    \item To the best of our knowledge, this is the first systematic study to leverage privileged knowledge distillation for time series forecasting. We propose TimeKD, an effective and efficient time series forecasting framework with calibrated language models.
    \item We develop a cross-modality teacher model comprising the calibrated language models and the subtractive cross attention mechanism, facilitating effective future time series representation extraction. 
    \item We propose the privileged knowledge distillation, which includes correlation and feature distillations leveraging privileged information, enables the student model to learn the teacher's behavior while minimizing the output discrepancy between them.
    \item We report on experiments on real data, offering evidence of the effectiveness, efficiency, and scalability of the proposed TimeKD. 
\end{itemize}

The remainder of this paper is structured as follows. Section~\ref{sec:related_work} surveys related work. Section~\ref{sec:problem} covers preliminary concepts and formalizes the problem of MTSF. We detail the TimeKD framework in Section~\ref{sec:method}, followed by the experimental study in Section~\ref{sec:exp}, and Section~\ref{sec:conclusion} concludes the paper.

\section{Related Work}
\label{sec:related_work}
We briefly review prior studies on LLMs-based time series forecasting and LLMs-based knowledge distillation.

\subsection{LLMs-based Time-Series Forecasting}
\re{Time series forecasting, which predicts events based on sequential temporal data, has gained increasing attention over the past decades~\cite{cirstea2021enhancenet,DBLP:journals/isci/JinLXSLH22,chenxi2021study}.}
Recently, LLMs-based time series forecasting methods~\cite{cao2023tempo, DBLP:conf/nips/ZhouNW0023, chang2023llm4ts, sun2023test, DBLP:journals/corr/abs-2406-01638} have attracted increasing interest due to the powerful knowledge transfer capability of LLMs and the increasing availability of time series data.
These models aim to harness the cross-domain knowledge and the transfer learning capabilities of LLMs for temporal correlation capturing. In particular, recent studies, e.g., TEMPO~\cite{cao2023tempo}, OFA~\cite{DBLP:conf/nips/ZhouNW0023}, LLM4TS~\cite{chang2023llm4ts}, and TEST~\cite{sun2023test} perform time series forecasting by fine-tuning. These models have integrated different mechanisms to adapt the LLMs to capture a variety of temporal semantics facilitating time series forecasting. Further, LLM-based methods utilize multiple data modalities to enhance time series forecasting, for example, introducing prompts with text as additional inputs enabling effective time series feature extraction. 

Existing multimodal LLMs for time series modeling methods can be divided into channel-independent time series forecasting models~\cite{jin2023time, liu2024unitime, xue2023promptcast, pan2024s, liu2024autotimes} and channel-dependent time series forecasting models~\cite{DBLP:journals/corr/abs-2406-01638, liu2024spatial}. 
Channel-independent time series forecasting methods~\cite{jin2023time, liu2024unitime, xue2023promptcast} often treat each time series variable separately. Time-LLM reprograms an LLM for time series forecasting where the backbone language model remains intact~\cite{jin2023time}. UniTime introduces a unified model that leverages natural language prompts and a Language-TS Transformer to align domain-specific characteristics for effective cross-domain time series forecasting~\cite{liu2024unitime}. 
However, these models often rely on channel independence, overlooking dependencies between multiple variables, resulting in longer training times and suboptimal performance. 
Channel-dependent time series forecasting methods~\cite{DBLP:journals/corr/abs-2406-01638} aims to extract effective features by learning correlations across multiple variables in time series based on LLMs. TimeCMA~\cite{DBLP:journals/corr/abs-2406-01638} focuses on handling data entanglement through cross-modality alignment. 
However, it remains a key problem to develop lightweight, efficient, and scalable LLMs-based time series forecasting methods. 

\subsection{LLMs-based Knowledge Distillation}


Knowledge distillation (KD) can generally be categorized into black-box distillation~\cite{DBLP:conf/ijcai/LiuHZLM24} and white-box distillation\cite{DBLP:conf/iclr/Gu0WH24}. In black-box KD, the student model has access only to the teacher model's predictions, while in white-box KD, the student can directly leverage the internal weights of the teacher model~\cite{li2024self}. With the rapid advancements in large language models (LLMs), black-box distillation KD has emerged as an effective technique to leverage cross-domain knowledge and mitigate the high computational costs. 
These LLM-based KD have been applied for various tasks, such as language generation~\cite{kang2024knowledge,zhao2024multistage}, graph learning~\cite{pan2024distilling}, and recommendation~\cite{li2023prompt}. 

Recently, LLMs-based knowledge distillation~\cite{DBLP:conf/ijcai/LiuHZLM24, liu2024taming} has been employed in time series.
For example,
AnomalyLLM~\cite{DBLP:conf/ijcai/LiuHZLM24}, introduces a knowledge distillation framework for time series anomaly detection, where the student model is trained to replicate the output features of an LLM. This approach enhances the training process by augmenting the time series data and inputting these samples into a partially frozen LLM. However, LLMs are not inherently designed for time series data, as they are pre-trained on extensive language corpora rather than numerical sequences. Moreover, the linear transformations involved in the anomaly detection would obscure information embedded in the teacher model's parameters. As a result, relying solely on the LLM’s final features for training the student model is insufficient to capture the knowledge required for downstream tasks.

\begin{figure}[t]
    \begin{centering}
\includegraphics[width=0.4\textwidth]{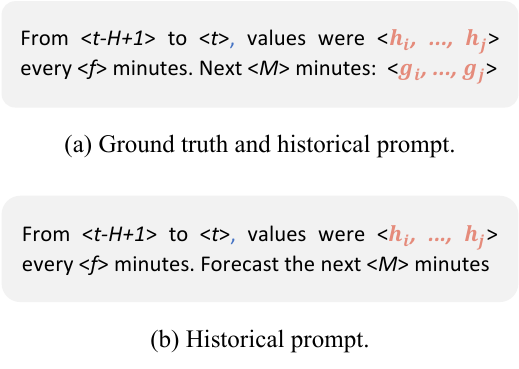}
        \caption{Examples of input prompts}
        \label{fig:prompt}
    \end{centering}
\end{figure}

\begin{figure*}[t]
    \begin{centering}	
    \includegraphics[width=\textwidth]{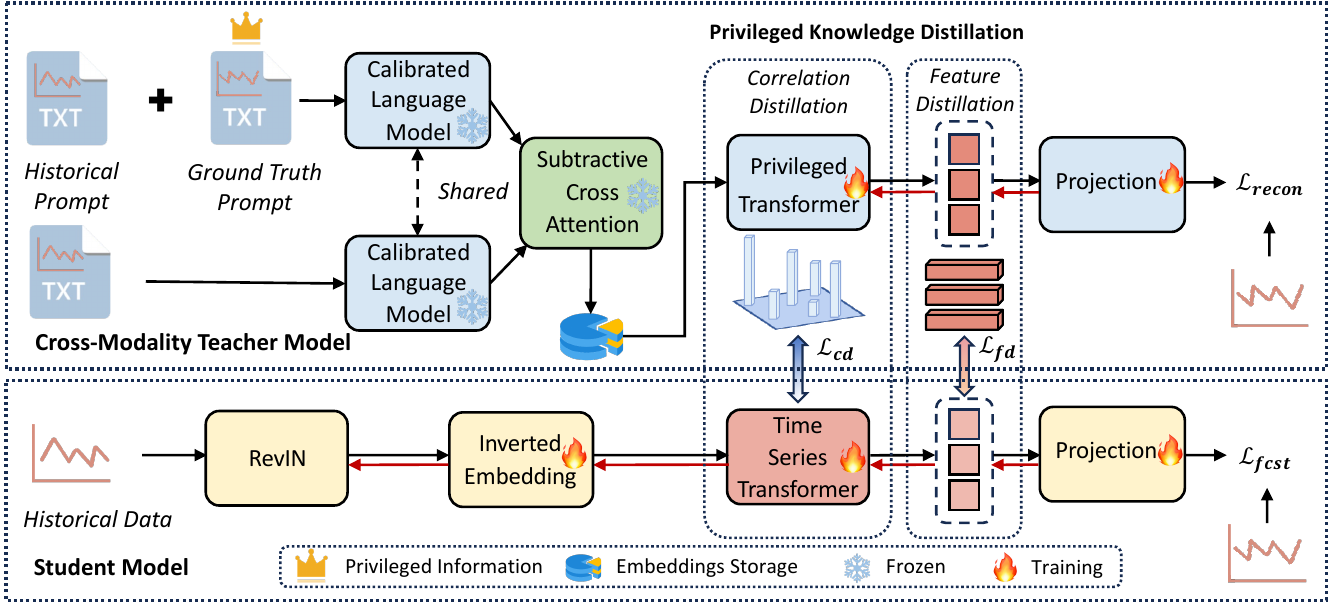}
    \caption{TimeKD Framework. 
    \textbf{Cross-Modality Teacher Model} processes textual prompts to reconstruct time series during the training stage. \textbf{Student Model} learned from the teacher model via privileged knowledge distillation for efficient forecasting. 
    }
    \label{fig:framework}
    \end{centering}
\end{figure*}

\section{Problem Statement}
\label{sec:problem}
We proceed to present the necessary preliminaries and then define the problem addressed.

\textbf{\textit{Definition 1 (Multivariate Time Series).}} The multivariate time series observations, denoted as $\mathbf{X} =\left\{\mathbf{x}_{1}, \ldots, \mathbf{x}_{|X|}\right\} \in \mathbb{R}^{|X| \times N}$, is a time ordered sequence of observations, where $N$ is the number of variables. Each observation $\mathbf{x}_i$ is a $N$-dimensional vector indicating $N$ features (e.g., load or temperature) at time step $i$. The term $v_i$ refers to the value of the time series at time step $i$. 
Notably, we define three types of multivariate time series: the historical time series $\mathbf{X}_H \in \mathbb{R}^{H \times N}$, the Ground truth $\mathbf{X}_G \in \mathbb{R}^{G \times N}$, and the observed time series $\mathbf{X}_O \in \mathbb{R}^{O \times N}$. The sequences $\mathbf{X}_H$ and $\mathbf{X}_G$ are employed during the training stage to develop a high-quality teacher model, while $\mathbf{X}_O$ is utilized in the test stage for forecasting.

\textbf{\textit{Definition 2 (Prompt).}} The prompt integrates time series values in the text format with a pre-defined template~\cite{xue2023promptcast,DBLP:journals/corr/abs-2406-01638}, word instructions about the time series forecasting task (please see Figure~\ref{fig:prompt}). Specifically, the historical time series $\mathbf{X}_H \in \mathbb{R}^{H \times N}$ are transformed into a set of textual prompt $\mathbf{P}_\text{HD} \in \mathbb{R}^{W_\text{HD} \times N}$, where $W_\text{HD}$ indicates the number of words in each historical data prompt, corresponding to the $H$ steps of time series. $\text{HD}$ denotes historical data.

\re{\textbf{Multivariate Time Series Forecasting.} Given a time series with $O$ observations $\mathbf{X}_O \in \mathbb{R}^{O \times N}$, the objective is to forecast future time series $\mathbf{X}_M \in \mathbb{R}^{M \times N}$ over $M$ time steps through a student model $f(\cdot)$ learned from a teacher model (i.e., LLMs) based on privileged knowledge distillation. Specifically, $f(\cdot)$ is trained using the historical time series $\mathbf{X}_H$ along with its textual prompt $\mathbf{P}_\text{HD}$, as well as ground truth $\mathbf{X}_G$ and its corresponding prompt $\mathbf{P}_\text{GT}$.}

\section{Methodology}
\label{sec:method}
In this section, we introduce TimeKD, an efficient multivariate \underline{time} series forecasting framework via calibrated language models with privileged \underline{k}nowledge \underline{d}istillation. We first give an overview of the framework and then provide specifics.

\subsection{Overall Framework}
\re{As illustrated in Figure~\ref{fig:framework}, TimeKD comprised a cross-modality teacher model and a student model with privileged knowledge distillation used to transfer knowledge from the teacher model and thereby train a powerful student model.}

\re{\textbf{Cross-Modality Teacher Model.} This model mainly consists of calibrated language models (CLMs), a subtractive cross attention (SCA), and a privileged Transformer $\mathit{PTEncoder(\cdot)}$ for the reconstruction task.
The ground truth $\mathbf{P}_{\text{GT}}$ and historical data $\mathbf{P}_{\text{HD}}$ prompts as privileged information, which are input to the CLMs respectively,
to facilitate effective future representation generation. The SCA is designed to remove textual information doped in the future time series representations. Then, we input these features into the $\mathit{PTEncoder(\cdot)}$, a lightweight Pre-LN Transformer encoder~\cite{DBLP:conf/icml/XiongYHZZXZLWL20}, to reconstruct the time series ground truth $\mathbf{X}_G$.}

\begin{itemize}
    \item \re{\textbf{Calibrated Language Models.} This module generates future time-series features through a combination of components: a tokenizer, layer normalizations, calibrated attention mechanisms, and feed-forward networks. The calibrated attention mechanisms are specifically designed to ensure modality consistency between textual and time-series tokens. The last token embeddings are extracted for efficient knowledge distillation.}

    \item \re{\textbf{Subtractive Cross Attention.} The module removes textual information doped in the last token embeddings, ensuring that the retained features are highly relevant for time series. These features are stored and utilized as privileged information for cultivating the student model, reducing the computational costs.}
\end{itemize}

\re{\textbf{Student Model.} This model processes historical data $\mathbf{X}_H$ through a reversible instance normalization layer (RevIN)~\cite{DBLP:conf/iclr/KimKTPCC22}, followed by an inverted embedding. The inverted embedding layer embeds the whole time series of each variate independently. Then, a time series Transformer $\mathit{TSTEncoder(\cdot)}$, which is a lightweight Transformer with the same structure as $\mathit{PTEncoder(\cdot)}$, processesl these embeddings to capture long-term temporal dependencies across multiple variables for forecasting the time series ground truth $\mathbf{X}_G$.}

\re{\textbf{Privileged Knowledge Distillation.} This module transfers the future representations from the teacher to the student model through two losses: correlation distillation loss $\mathcal L_\text{cd}$ and feature distillation loss $\mathcal L_\text{fd}$. The $\mathcal L_\text{cd}$ aligns the correlations between the Transformers from the teacher and the student model, leveraging the student model to imitate the teacher's behavior. Meanwhile, the $\mathcal L_\text{fd}$ minimizes the output discrepancy between teacher and student models.}

\re{Finally, the observed time series $\mathbf{X}_O$ to used to predict the future time series $\mathbf{X}_M$ via a well-learned student model.}

\subsection{Cross-Modality Teacher Model}
We train an LLM-based cross-modality teacher model to reconstruct the time series ground truth for learning high-quality future representations.

\subsubsection{Calibrated Language Models}

\begin{figure}[t]
    \begin{centering}	
    \includegraphics[width=0.4\textwidth]{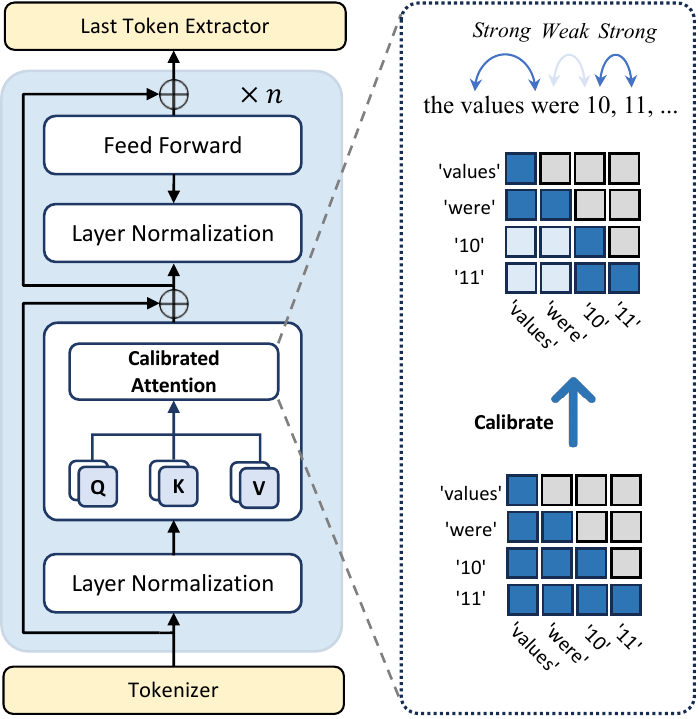}
        \caption{An example of calibrated attention score in the CLMs.}
        \label{fig:clm}
    \end{centering}
\end{figure}

Calibrated language models (CLMs) modify the masked multi-self attention in large language models (LLMs) based on the cross-modality and intra-modality correlations of prompt tokens. The CLMs are composed of a tokenizer, layer normalization layers, calibrated attention mechanisms, feed-forward networks, and a last token extractor, as shown in Figure~\ref{fig:clm}.

Given the historical data prompt $\mathbf{P}_\text{HD}=\left\{\mathbf{p}_1, \ldots, \mathbf{p}_{W_\text{HD}}\right\} \in \mathbb{R}^{W_\text{HD} \times N}$ and ground truth prompt $\mathbf{P}_\text{GT}=\left\{\mathbf{p}_1, \ldots, \mathbf{p}_{W_\text{GT}}\right\} \in \mathbb{R}^{{W_\text{GT}} \times N}$, where $W_\text{HD}$ and $W_\text{GT}$ are the words in each prompt, and $W_\text{HD} < W_\text{GT}$. We first input them into the CLMs separately to get the prompt embeddings. The tokenizer is responsible for converting these prompts into a series of token IDs $\mathbf{I}_\text{GT} \in \mathbb{R}^{S_\text{HD} \times N}$ and $\mathbf{I}_\text{HD} \in \mathbb{R}^{S_\text{GT} \times N}$, where $S_\text{HD}$ and $S_\text{GT}$ represents the number of token ID. 

Subsequently, these tokenized prompt representations are processed by the CLMs to generate contextualized embeddings. This involves a series of transformations, including calibrated attention, layer normalization, and feed-forward operations, which progressively refine the token representations across multiple layers:
\begin{align}
\overline{\boldsymbol{\mathcal{I}}}_\text{GT}^{i} &= \mathit{CalAtt}(\mathit{LN}(\mathbf{I}^{i}_\text{GT})) + \mathbf{I}^{i}_\text{GT}, \label{eq:CA_ln} \\
\boldsymbol{\mathcal{I}}_\text{GT}^{i+1} &= \mathit{FFN}(\mathit{LN}(\overline{\boldsymbol{\mathcal{I}}}_\text{GT}^{i})) + \overline{\boldsymbol{\mathcal{I}}}_\text{GT}^{i}, \label{eq:ffn_ln}
\end{align}
where $\overline{\boldsymbol{\mathcal{I}}}_\text{GT}^{i} \in \mathbb{R}^{S_\text{HD} \times N \times D}$ represents the intermediate representation of the $i_{\text{th}}$ layer after applying the $\mathit{CalAtt}(\cdot)$ and the $\mathit{LN}(\cdot)$. $D$ denotes the hidden dimension of the language model.
$\mathbf{I}^{0}_\text{GT} =\left[\mathbf{I}_\text{GT} + \mathbf{PE}\right]$
, where $\mathbf{PE}$ represents the learnable positional encoding.

Then, we design the calibrated attention mechanism $\mathit{CalAtt}$ to enhance the masked multi-self attention (MMSA) within LLMs for processing multi-modality data, such as time series and text. Traditional MMSA often struggles to distinguish the significance of cross-modality and intra-modality interactions, resulting in data entanglement issues~\cite{DBLP:journals/corr/abs-2406-01638}. For instance, as shown in Figure~\ref{fig:clm}, the original attention mask's attention scores (bottom) display a uniform distribution across the lower triangular region. In contrast, the calibrated attention mechanism (top) strengthens intra-modality interactions by reducing the weights of cross-modality interactions (e.g., between the time series token '10' and the text token 'were'). The formulation of $\mathit{CalAtt}$ is as follows.
\begin{align}
&\mathit{CalAtt}(\tilde{\mathbf{I}}^{i}_\text{GT}) = \xi_o \left(\mathit{Att}(\xi_q \tilde{\mathbf{I}}^{i}_\text{GT}, \xi_k \tilde{\mathbf{I}}^{i}_\text{GT}, \xi_v \tilde{\mathbf{I}}^{i}_\text{GT}) \right), \\
&\mathit{Att}(\tilde{\mathbf{I}}^{i}_\text{GT}, \tilde{\mathbf{I}}^{i}_\text{GT}, \tilde{\mathbf{I}}^{i}_\text{GT}) = \operatorname{softmax}\left(\frac{\tilde{\mathbf{I}}^{i}_\text{GT}\tilde{\mathbf{I}}^{i}_\text{GT}\top}{\sqrt{d_k}} + \text{Mask}\right) \tilde{\mathbf{I}}^{i}_\text{GT}, \\
&\text{Mask}[i, j] = 
\begin{cases}
-\Delta,  & \text{if tokens } i \text{ and } j \text{ are cross-modality}, \\
0, & \text{if tokens } i, j \text{ are intra-modality},
\end{cases}
\end{align}
where $\tilde{\mathbf{I}}^{i}_\text{GT}$ is the output of $\mathbf{I}^{i}_\text{GT}$ after passing through the first $\mathit{LN}$. $\mathit{Att}(\tilde{\mathbf{I}}^{i}_\text{GT}, \tilde{\mathbf{I}}^{i}_\text{GT}, \tilde{\mathbf{I}}^{i}_\text{GT})$ represents the attention mechanism that computes the attention-weighted combination of values $\tilde{\mathbf{I}}^{i}_\text{GT}$ using the query $\tilde{\mathbf{I}}^{i}_\text{GT}$ and key $\tilde{\mathbf{I}}^{i}_\text{GT}$, which are derived from the input $\tilde{\mathbf{I}}^{i}_\text{GT}$ through the linear transformations $\xi_q$, $\xi_k$, and $\xi_v$, respectively. The term $d_k$ represents the dimensionality of the key vectors, and it is used to scale the dot-product attention scores for numerical stability. The term $\text{Mask}$ adjusts the attention scores by penalizing cross-modality interactions. 

The $\mathit{LN}$ and $\mathit{FFN}$ in the CLMs are defined as follows:
\begin{align}
& \mathit{LN}\left(\mathbf{I}^{i}_\text{GT}\right) = \gamma \odot \frac{\mathbf{I}^{i}_\text{GT} - \mu}{\sigma} + \beta, \label{eq:layer_norm} \\
& \mathit{FFN}(\hat{\mathbf{I}}^{i}_\text{GT}) = \max(0, \mathbf W_1\hat{\mathbf{I}}^{i}_{\text{GT}} + \mathbf b_1)\mathbf W_2 + \mathbf b_2, \label{eq:feed_forward}
\end{align}
where $\hat{\mathbf{I}}^{i}_\text{GT}$ is the output of $\mathbf{I}^{i}_\text{GT}$ after the second $\mathit{LN}$. $\gamma$ and $\beta$ are learnable scaling and translation parameters. $\mu$ and $\sigma$ represent the mean and standard deviation, respectively. $\odot$ denotes element-wise multiplication.

The outputs of the CLMs are denoted as $\boldsymbol{\mathcal{I}}_\text{GT} \in \mathbb{R}^{S_\text{GT} \times N \times D}$ and $\boldsymbol{\mathcal{I}}_{HD} \in \mathbb{R}^{S_\text{HD} \times N \times D}$. We then extract the last token embeddings from these outputs, which is motivated by the observation that the last token in a prompt encapsulates the most comprehensive knowledge due to the masked attention mechanism in LLMs. Specifically, the representation of the last token at a given position is influenced solely by the representations of its preceding tokens. As illustrated in Figure~\ref{fig:clm}, assuming '11' is the last token in a prompt, it can attend to the token '10' (highlighted by the blue cubes). However, token '10' cannot attend to '11' due to the masking mechanism.

To leverage this property and reduce computational overhead for efficient knowledge distillation, we extract the last token embeddings: $\mathbf{L}_\text{GT}= \left\{\mathbf{l}_1, \ldots, \mathbf{l}_{N}\right\} \in \mathbb{R}^{N \times D}$ from the $\boldsymbol{\mathcal{I}}_\text{GT} \in \mathbb{R}^{S_\text{HD} \times N \times D}$ and $\mathbf{L}_\text{HD}= \left\{\mathbf{l}_1, \ldots, \mathbf{l}_{N}\right\} \in \mathbb{R}^{N \times D}$ from the $\boldsymbol{\mathcal{I}}_\text{HD} \in \mathbb{R}^{S_\text{GT} \times N \times D}$. This strategy ensures computational efficiency while preserving the essential information for subsequent distillation. 

\subsubsection{Subtractive Cross Attention}
\begin{figure}[t]
    \begin{centering}	
    \includegraphics[width=0.48\textwidth]{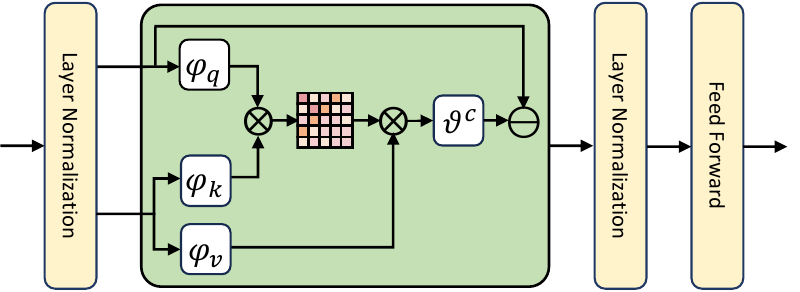}
        \caption{Subtractive cross attention.}
        \label{fig:sca}
    \end{centering}
\end{figure}

\re{We design a subtractive cross-attention (SCA) mechanism to eliminate textual information embedded in the last token representations, ensuring that the retained embedding remains highly relevant to time series forecasting.

In SCA, we begin by applying layer normalization and projection functions, denoted as $\varphi_q$, $\varphi_k$, and $\varphi_v$, to the projected embeddings of the ground truth $\mathbf{L}_\text{GT}$ and historical data $\mathbf{L}_\text{HD}$. We then compute the channel-wise similarity matrix $\mathbf{M}_C \in \mathbb{R}^{D \times D}$ via matrix multiplication, followed by a softmax operation:
\begin{equation}
\mathbf{M}_C = \operatorname{softmax}\left(\mathit{LN}(\varphi_q(\mathbf{L}_\text{GT})) \otimes \mathit{LN}\varphi_k(\mathbf{L}_\text{HD})\right),
\end{equation}
where $\otimes$ denotes matrix multiplication.

Next, we perform channel-wise feature aggregation by multiplying $\varphi_v(\mathbf{L}_\text{HD})$ with $\mathbf{M}_C$. The refined ground truth prompt embedding, $\overline{\mathbf{L}}_\text{GT}$, is then derived by subtracting this intersection from both $\mathbf{L}_\text{HD}$ and $\mathbf{L}_\text{GT}$, followed by layer normalization and a feed-forward layer:
\begin{equation}
\overline{\mathbf{L}}_\text{GT} = \mathit{FFN}(\mathit{LN}(\mathbf{L}_\text{GT} \ominus \vartheta^c\left(\varphi_v\left(\mathbf{L}_\text{HD}\right) \otimes \mathbf{M}_C\right))),
\end{equation}
where $\overline{\mathbf{L}}_\text{GT} \in \mathbb{R}^{N \times D}$ represents the refined ground truth embedding. Here, $\vartheta^c$ is a linear layer, and $\ominus$ denotes the subtraction operation. 

The SCA refines the ground truth prompt embedding, preserving privileged knowledge for further distillation. Additionally, to avoid repetitive processing with the frozen CLMs, we store the subtracted embeddings for efficient reconstruction.}

\subsubsection{Cross-Modality Reconstruction}
This task leverages a Transformer Encoder $\mathit{PTEncoder(\cdot)}$ to reconstruct the ground truth in the time series modality using $\overline{\mathbf{L}}_\text{GT}$ from the textual modality.

Inside $\mathit{PTEncoder(\cdot)}$, the $\overline{\mathbf{L}}_\text{GT}$ first undergo layer normalization $\mathit{LN}(\cdot)$ at the $i_{\text{th}}$ layer:
\begin{align}
\widetilde{\mathbf{L}}_\text{GT}^{i} &= \mathit{LN}(\overline{\mathbf{L}}_\text{GT}),
\end{align}
where $\widetilde{\mathbf{L}}_\text{GT}^{i}$ represents the intermediate embedding after normalization. The structure of $\mathit{LN}(\cdot)$ is the same as in Equation~(\ref{eq:layer_norm}).

The normalized embeddings $\widetilde{\mathbf{L}}_\text{GT}^{i}$ are then passed through a multi-head attention layer within $\mathit{PTEncoder(\cdot)}$, denoted as $\mathit{PTAtt}(\cdot)$. The output, $\widecheck{\mathbf{L}}_\text{GT}^{i}$, is combined with the input through a residual connection:
\begin{align}
\widecheck{\mathbf{L}}_\text{GT}^{i} &= \mathit{PTAtt}(\widetilde{\mathbf{L}}_\text{GT}^{i}) + \overline{\mathbf{L}}_\text{GT}^{i}, \\
\mathit{PTAtt}(\widetilde{\mathbf{L}}_\text{GT}^{i}) &= \zeta_o \big(\mathit{Att}(\zeta_q \widetilde{\mathbf{L}}_\text{GT}^{i}, \zeta_k \widetilde{\mathbf{L}}_\text{GT}^{i}, \zeta_v \widetilde{\mathbf{L}}_\text{GT}^{i}) \big), \\
\mathit{Att}(\widetilde{\mathbf{L}}_\text{GT}^{i}) &= \mathit{softmax}\left(\frac{\widetilde{\mathbf{L}}_\text{GT}^{i} \widetilde{\mathbf{L}}^{i\top}_\text{GT}}{\sqrt{d_k}}\right) \widetilde{\mathbf{L}}_\text{GT}^{i},
\end{align}
where $\zeta_o$, $\zeta_q$, $\zeta_k$, and $\zeta_v$ are learnable linear projections. The attention computes dependencies across feature dimensions.

The output $\widecheck{\mathbf{L}}_\text{GT}^{i}$ is then normalized again through another $\mathit{LN}(\cdot)$, followed by a feed-forward network $\mathit{FFN}(\cdot)$. The result is combined with the input via a residual connection:
\begin{align}
\mathbf{E}_\text{GT}^{i+1} &= \mathit{FFN}(\mathit{LN}(\widecheck{\mathbf{L}}_\text{GT}^{i})) + \widecheck{\mathbf{L}}_\text{GT}^{i},
\end{align}
where $\mathbf{E}_\text{GT}^{i+1} \in \mathbb{R}^{D \times N}$ represents the output of $\mathit{PTEncoder(\cdot)}$, denoted as $\mathbf{E}_\text{GT}$ for simplify, and the structure of $\mathit{FFN}(\cdot)$ is identical to Equation~(\ref{eq:feed_forward}).

The reconstructed time series is then generated by applying a projection function to $\mathbf{E}_\text{GT}$:
\begin{equation}
\widehat{\mathbf{X}}_G = \mathbf{W}_l \mathbf{E}_\text{GT} + \mathbf{b}_l,
\end{equation}
where $\widehat{\mathbf{X}}_G \in \mathbb{R}^{G \times N}$ is the reconstructed time series, and $\mathbf{W}_l$ and $\mathbf{b}_l$ are learnable parameters.

The reconstruction loss is defined using SmoothL1 loss ($\mathit{SL1}$):
\begin{equation}
\mathcal{L}_{\text{recon}} = \frac{1}{\mathrm{G}} \sum_{G=1}^{\mathrm{G}} \mathit{SL1}(\mathbf{\hat{X}}_G - \mathbf{X}_G),
\end{equation}
\begin{equation}
\mathit{SL1}(\mathbf{\hat{X}}_G - \mathbf{X}_G) = 
\begin{cases} 
0.5(\mathbf{\hat{X}}_G - \mathbf{X}_G)^2, & \text{if } |\mathbf{\hat{X}}_G - \mathbf{X}_G| < 1, \\
|\mathbf{\hat{X}}_G - \mathbf{X}_G| - 0.5, & \text{otherwise}, \label{eq:sl1}
\end{cases}
\end{equation}
where $\mathrm{L}$ is the total number of reconstruction samples. The SmoothL1 loss ensures robustness to outliers while maintaining sensitivity to smaller errors.

\begin{algorithm}[t]
\caption{Cross-Modality Teacher Model Training}
\label{alg:teacher_training}
\textbf{Input:} Ground truth with historical prompts $\mathbf{P}_\text{GT}$ and historical prompts $\mathbf{P}_\text{HD}$ \\
\textbf{Output:} Privileged Transformer attentions $\mathbf{A}_\text{PE}$, privileged embeddings $\mathbf{E}_\text{GT}$, and reconstructed time series $\mathbf{\hat{X}}_L$
\begin{algorithmic}[1]
\While{not converged}
    \State $\mathbf{L}_\text{HD}, \mathbf{L}_\text{GT} \gets \mathit{CLM}(\mathbf{P}_\text{HD}), \mathit{CLM}(\mathbf{P}_\text{GT})$
    \State $\overline{\mathbf{L}}_\text{GT} \gets SCA(\mathbf{L}_\text{GT},\mathbf{L}_\text{HD})$
    \State $\mathbf{E}_\text{GT}, \mathbf{A}_\text{PE} \gets \mathit{PTEncoder}(\overline{\mathbf{L}}_\text{GT})$
    \State $\mathbf{\hat{X}}_G \gets \mathbf{W}_l \mathbf{E}_\text{GT} + \mathbf{b}_l$
    \State $\mathcal{L}_{\text{recon}} \gets \frac{1}{\mathrm{L}} \sum_{L=1}^{\mathrm{L}} \mathit{SL1}(\mathbf{\hat{X}}_G - \mathbf{X}_G)$
    \State Update model parameters using $\mathcal{L}_{\text{recon}}$
\EndWhile
\State \textbf{Return:} $\mathbf{A}_\text{PE}, \mathbf{E}_\text{GT}, \mathbf{\hat{X}}_G$
\end{algorithmic}
\end{algorithm}

\subsection{Student Model}
The student model processes historical time series through the RevIN, an inverted embedding layer, a time series Transformer, and a projection layer. Given historical data $\mathbf{X}_H \in \mathbb{R}^{H \times N}$, the inverted embedding transforms $\mathbf{X}_H$ into learnable matrices $\mathbf{I}_H \in \mathbb{R}^{H \times N}$ to capture temporal dependencies across multiple variables~\cite{DBLP:journals/corr/abs-2310-06625}. The $\mathbf{X}_O$ is first normalized via RevIN to mitigate distribution shifts. The normalized data is then embedded as follows:
\begin{equation}
\mathbf{I}_H = \mathbf{W}_i \mathbf{X}_H + \mathbf{b}_i,
\end{equation}
where $\mathbf{I}_H = \left\{\mathbf{i}_1, \ldots, \mathbf{i}_T\right\} \in \mathbb{R}^{D \times N}$ is the output of the inverted embedding layer. $C$ denotes the hidden dimension of the embeddings, while $\mathbf{W}_i$ and $\mathbf{b}_i$ are learnable parameters.

The embeddings $\mathbf{I}_H$ are then passed into a Transformer encoder $\mathit{TSTEncoder(\cdot)}$, which models the temporal dependencies across multiple variables. The time series encoder consists of multiple layers, each with the following pipeline:
\begin{align}
\widetilde{\mathbf{T}}_H^{i} &= \mathit{LN}(\mathbf{I}_H), \\
\overline{\mathbf{T}}_H^{i} &= \mathit{TSTAtt}(\widetilde{\mathbf{T}}_H^{i}) + \mathbf{T}^{i}_H, \\
\overline{\mathbf{T}}_{H}^{i+1} &= \mathit{FFN}(\mathit{LN}(\grave{\mathbf{T}}^{i}_H)) + \overline{\mathbf{T}}^{i}_H,
\end{align}
where $\widetilde{\mathbf{T}}_H^i$ represents the intermediate embedding after layer normalization and $\overline{\mathbf{T}}_H^i$ is the output combined with the input via a residual connection.

Multi-head self-attention mechanism in the encoder $\mathit{TSTAtt}(\cdot)$ computes the attention-weighted representations:
\begin{align}
\mathit{TSTAtt}(\widetilde{\mathbf{T}}_H^{i}) &= \eta_o (\mathit{Att}(\eta_q \widetilde{\mathbf{T}}_H^{i}, \eta_k \widetilde{\mathbf{T}}_H^{i}, \eta_v \widetilde{\mathbf{T}}_H^{i})), \\
\mathit{Att}(\widetilde{\mathbf{T}}_H^{i}) &= \mathit{softmax}\left(\frac{\widetilde{\mathbf{T}}_H^{i} \widetilde{\mathbf{H}}^{i\top}_L}{\sqrt{d_k}}\right) \widetilde{\mathbf{T}}_H^{i},
\end{align}
where $\eta_o$, $\eta_q$, $\eta_k$, and $\eta_v$ are learnable projection matrices, and $d_k$ is the dimensionality of the key vectors. 

The final output $\overline{\mathbf{T}}_H \in \mathbb{R}^{N \times D}$ from the $\mathit{TSTEncoder(\cdot)}$ enriched temporal and cross-variable features, which are used for downstream tasks.

\subsection{Privileged Knowledge Distillation}

Privileged knowledge distillation emphasizes correlation and feature distillation to transfer the privileged representations from the high-quality teacher to a lightweight student model.

\subsubsection{Correlation Distillation}
Correlation distillation aligns the attention maps between the teacher network's privileged Transformer, $\mathit{PTEncoder(\cdot)}$, and the student's time series Transformer, $\mathit{TSTEncoder(\cdot)}$, leveraging the student model to imitate the teacher’s behavior. The teacher's calibrated attention maps guide the student's attention distributions to preserve relational dependencies across features. Specifically, the attention matrices $\mathbf{A}_\text{PE} \in \mathbb{R}^{N \times N}$ from $\mathit{PTEncoder(\cdot)}$ and $\mathbf{A}_\text{TSE} \in \mathbb{R}^{N \times N}$ from $\mathit{TSTEncoder(\cdot)}$ are averaged across all heads of the last encoder layer to form unified representations. The correlation distillation loss is defined as:
\begin{equation}
\mathcal{L}_\text{{cd}} = \frac{1}{|\mathbf{A}_\text{PE}|} \sum_{i} \mathit{SL1}\left(\mathbf{A}_\text{PE}^i - \mathbf{A}_\text{TSE}^i\right),
\end{equation}
where $\mathit{SL1}(\cdot)$, defined in Equation~(\ref{eq:sl1}), ensures that the student replicates the teacher's contextual understanding of feature features while reducing sensitivity to outliers.

\subsubsection{Feature Distillation}
Feature distillation aligns the embedding spaces of the teacher and student models. The teacher generates privileged embeddings, $\mathbf{E}_\text{GT} \in \mathbb{R}^{N \times D}$, from $\mathit{PTEncoder(\cdot)}$, which the student replicates through its time series Transformer, producing embeddings $\overline{\mathbf{T}}_H \in \mathbb{R}^{N \times D}$ from $\mathit{TSTEncoder(\cdot)}$. The feature distillation loss is similarly implemented as a Smooth L1 Loss:
\begin{equation}
\mathcal{L}_\text{fd} = \frac{1}{|\mathbf{E}_\text{GT}|} \sum_{i} \mathit{SL1}\left(\mathbf{E}_\text{GT}^i - \overline{\mathbf{T}}_H^i\right).
\end{equation}

\noindent This process ensures that the student effectively captures the enriched features learned by the teacher, enabling robust and accurate knowledge transfer.

The overall distillation loss combines correlation distillation loss $\mathcal{L}_\text{cd}$ and feature distillation $\mathcal{L}_\text{fd}$ loss to guide the student's learning process. The total knowledge distillation loss is defined as:
\begin{equation} 
\mathcal{L}_\text{PKD} = \lambda_\text{c} \mathcal{L}_\text{CD} + \lambda_\text{e} \mathcal{L}_\text{FD}, \end{equation}
where $\lambda_\text{c}$ and $\lambda_\text{e}$ are the hyperparameters that balance the contributions of the correlation distillation loss $\mathcal{L}_\text{CD}$ and feature distillation $\mathcal{L}_\text{FD}$ loss. 

\begin{algorithm}[t]
\caption{Privileged Knowledge Distillation}
\label{alg:privileged_kd}
\textbf{Input:} Privileged embeddings $\mathbf{E}_\text{GT}$, privileged Transformer attentions $\mathbf{A}_\text{PE}$, time series embeddings $\overline{\mathbf{T}}_H$, and time series Transformer attentions $\mathbf{A}_\text{TSE}$ \\
\textbf{Output:} Distilled student model $\mathit{DST(\cdot)}$
\begin{algorithmic}[1]
\While{not converged}
    \State $\mathcal{L}_{\text{cd}} \gets \frac{1}{|\mathbf{A}_\text{PE}|} \sum_{i} \mathit{SL1}(\mathbf{A}_\text{PE}^i - \mathbf{A}_\text{TSE}^i)$
    \State $\mathcal{L}_{\text{fd}} \gets \frac{1}{|\mathbf{E}_\text{GT}|} \sum_{i} \mathit{SL1}(\mathbf{E}_\text{GT}^i - \overline{\mathbf{T}}_H^i)$
    \State $\mathcal{L}_{\text{pkd}} \gets \lambda_\text{c} \mathcal{L}_{\text{cd}} + \lambda_\text{f} \mathcal{L}_{\text{fd}}$
    \State Update student model $\mathit{DST(\cdot)}$ parameters using $\mathcal{L}_{\text{PKD}}$
\EndWhile
\State \textbf{Return:} Distilled student model $\mathit{DST(\cdot)}$
\end{algorithmic}
\end{algorithm}

\subsection{Time Series Forecasting}
The time series forecasting module leverages the well-learned student model for efficient forecasting.

During the test process, only the student model is employed for inference. Specifically,
$\overline{\mathbf{H}}_O$ is the time series embedding from the distilled student model $\mathit{DST(\cdot)}$. Then, the $\overline{\mathbf{H}}_O$ is input into a projection function for future prediction. It can be formulated as follows.
\begin{equation}
    \overline{\mathbf{H}}_O = \mathit DST(\mathbf{X}_O),
\end{equation}
\begin{equation}
    \widehat{\mathbf{X}}_M = \mathbf{W}_t \overline{\mathbf{H}}_O + \mathbf{b}_t,
\end{equation}
where $\widehat{\mathbf{X}}_M \in \mathbb{R}^{M \times N}$ represents the projected forecasts. $\mathbf{W}_t$ and $\mathbf{b}_t$ are the learnable parameters. Finally, the output $\widehat{\mathbf{X}}_M$ is normalized. The forecasting loss is SmoothL1 loss:
\begin{equation}
\mathcal{L}_{\text{fcst}} = \frac{1}{\mathrm{M}} \sum_{M=1}^{\mathrm{M}} \mathit{SL1}(\mathbf{\hat{X}}_M - \mathbf{X}_M,),
\end{equation}
where $\mathrm{M}$ is the number of forecasting sample sizes.

The loss function of the proposed TimeKD consists of four parts: a reconstruction loss $\mathcal{L}_{\text{recon}}$, a correlation distillation loss $\mathcal{L}_{\text{cd}}$, a feature distillation loss $\mathcal{L}_{\text{fd}}$, and a forecasting loss $\mathcal{L}_{\text{fcst}}$. We combine them and obtain the overall loss as follows.

\begin{equation}
    \mathcal{L}_{\text{Task}} = \lambda_\text{r} \mathcal{L}_{\text{Recon}} + \lambda_\text{p} \mathcal{L}_{\text{PKD}} + \lambda_\text{f} \mathcal{L}_{\text{Fcst}},
\end{equation}
where $\lambda_{r},~\lambda_{p}$ and $\lambda_{f}$ hyperparameters that balance 
\section{Experimental Evaluation}
\label{sec:exp}

\subsection{Experiment Setup}
\subsubsection{Datasets}

We conduct experiments with eight widely-used time series datasets for time series forecasting. Specifically, ETTm1, ETTm2, ETTh1, ETTh2~\cite{Zeng2022AreTE}, Exchange~\cite{DBLP:conf/sigir/LaiCYL18} Weather~\cite{DBLP:conf/nips/WuXWL21} are utilized for long-term forecasting, while PEMS04, and PEMS08~\cite{DBLP:journals/corr/abs-2310-06625} are employed for short-term forecasting. These datasets span four application domains: electricity, economy, weather, and traffic. Additionally, textual prompts are designed on the time series data.

\begin{itemize}
    \item \textbf{ETT} includes hourly-level datasets (ETTh1 and ETTh2) and 15-minute-level datasets (ETTm1 and ETTm2). Each dataset includes $7$ oil and load features of electricity transformers between July 2016 and July 2018.
    \item \textbf{Weather} comprises 21 weather indicators, such as air temperature and humidity, collected in Germany. The data is recorded every 10 minutes. 
    \item \textbf{Exchange} consists of daily exchange rates from eight countries, including Australia, the United Kingdom, Canada, Switzerland, China, Japan, New Zealand, and Singapore, covering the period from 1990 to 2016.
    \item \re{\textbf{PEMS} comprises public traffic network data from California, collected in 5-minute intervals. We utiliz two public subsets, PEMS04 and PEMS08, as adopted in existing studies~\cite{DBLP:journals/corr/abs-2310-06625,hu2025timefilter}.}
\end{itemize}

\begin{table*}[t]
\centering
\caption{Long-term Forecasting performance comparisons. The input length is 96 for all datasets.
}
\resizebox{0.95\textwidth}{!}{
\begin{tabular}{cc|cc|cc|cc|cc|cc|cc|cc}
\toprule
\multicolumn{2}{c|}{\multirow{2}{*}{Datasets}}                           & \multicolumn{2}{c|}{TimeKD}     & \multicolumn{2}{c|}{TimeCMA} & \multicolumn{2}{c|}{TimeLLM} & \multicolumn{2}{c|}{UniTime} & \multicolumn{2}{c|}{OFA} & \multicolumn{2}{c|}{iTransformer} & \multicolumn{2}{c}{PatchTST} \\ \cmidrule{3-16}                     &   & MSE            & MAE            & MSE           & MAE          & MSE           & MAE          & MSE           & MAE          & MSE            & MAE     & MSE             & MAE             & MSE              & MAE       \\ \midrule
\multicolumn{1}{c|}{\multirow{6}{*}{ETTm1}}    & 24  & \re{\textbf{0.202}} & \re{\textbf{0.258}} & {\ul 0.208}   & {\ul 0.281}  & 0.263         & 0.232        & 0.228         & 0.301        & 0.221          & 0.293   & 0.223           & 0.296           & 0.210            & 0.283     \\
\multicolumn{1}{c|}{}                          & 36  & \re{\textbf{0.259}} & \re{\textbf{0.302}} & {\ul 0.267}   & {\ul 0.319}  & 0.296         & 0.344        & 0.273         & 0.330        & 0.269          & 0.324   & 0.274           & 0.331           & 0.272            & 0.325     \\
\multicolumn{1}{c|}{}                          & 48  & \re{\textbf{0.272}} & \re{\textbf{0.323}} & {\ul 0.282}   & {\ul 0.329}  & 0.295         & 0.343        & 0.298         & 0.346        & 0.292          & 0.338   & 0.312           & 0.356           & 0.286            & 0.335     \\
\multicolumn{1}{c|}{}                          & 96  & \re{\textbf{0.305}} & \re{\textbf{0.342}} & {\ul 0.312}   & {\ul 0.351}  & 0.359         & 0.381        & 0.328         & 0.367        & 0.326          & 0.362   & 0.334           & 0.373           & 0.320            & 0.359     \\
\multicolumn{1}{c|}{}                          & 192 & \re{\textbf{0.355}} & \re{\textbf{0.367}} & {\ul 0.361}   & {\ul 0.378}  & 0.383         & 0.393        & 0.368         & 0.387        & 0.367          & 0.382   & 0.377           & 0.391           & 0.362            & 0.381     \\ \cmidrule{2-16} 
\multicolumn{1}{c|}{}                          & Avg & \re{\textbf{0.279}} & \re{\textbf{0.318}} & {\ul 0.286}   & {\ul 0.331}  & 0.319         & 0.357        & 0.299         & 0.346        & 0.295          & 0.340   & 0.304           & 0.349           & 0.290            & 0.337     \\ \midrule
\multicolumn{1}{c|}{\multirow{6}{*}{ETTm2}}    & 24  & \re{\textbf{0.083}} & \re{\textbf{0.184}} & {\ul 0.092}   & {\ul 0.190}  & 0.102         & 0.198        & 0.105         & 0.201        & 0.103          & 0.201   & 0.104           & 0.200           & 0.099            & 0.193     \\
\multicolumn{1}{c|}{}                          & 36  & \re{\textbf{0.109}} & \re{\textbf{0.204}} & {\ul 0.116}   & {\ul 0.211}  & 0.122         & 0.218        & 0.124         & 0.221        & 0.124          & 0.222   & 0.123           & 0.219           & 0.119            & 0.215     \\
\multicolumn{1}{c|}{}                          & 48  & \re{\textbf{0.125}} & \re{\textbf{0.214}} & {\ul 0.130}   & {\ul 0.225}  & 0.137         & 0.230        & 0.140         & 0.234        & 0.138          & 0.236   & 0.139           & 0.234           & 0.134            & 0.229     \\
\multicolumn{1}{c|}{}                          & 96  & \re{\textbf{0.170}} & \re{\textbf{0.251}} & {\ul 0.173}   & {\ul 0.258}  & 0.193         & 0.280        & 0.181         & 0.263        & 0.176          & 0.261   & 0.180           & 0.264           & 0.177            & 0.260     \\
\multicolumn{1}{c|}{}                          & 192 & \re{\textbf{0.225}} & \re{\textbf{0.296}} & {\ul 0.238}   & {\ul 0.301}  & 0.257         & 0.318        & 0.248         & 0.308        & 0.243          & 0.305   & 0.250           & 0.309           & 0.246            & 0.305     \\ \cmidrule{2-16} 
\multicolumn{1}{c|}{}                          & Avg & \re{\textbf{0.142}} & \re{\textbf{0.230}} & {\ul 0.150}   & {\ul 0.237}  & 0.162         & 0.249        & 0.160         & 0.245        & 0.157          & 0.245   & 0.159           & 0.245           & 0.155            & 0.240     \\ \midrule
\multicolumn{1}{c|}{\multirow{6}{*}{ETTh1}}    & 24  & \re{\textbf{0.290}} & \re{\textbf{0.346}} &  0.296   &  0.352  & 0.325         & 0.370        & 0.359         & 0.390        & {\ul0.295}          &{\ul 0.351}   & 0.304           & 0.359           & 0.321            & 0.365     \\
\multicolumn{1}{c|}{}                          & 36  & \re{\textbf{0.314}} & \re{\textbf{0.360}} & {\ul 0.318}   & {\ul 0.363}  & 0.348         & 0.383        & 0.380          & 0.401        & {\ul0.318}          & 0.364   & 0.328           & 0.372           & 0.343            & 0.377     \\
\multicolumn{1}{c|}{}                          & 48  & \re{\textbf{0.325}} & \re{\textbf{0.368}} &  0.340   & 0.375  & 0.358         & 0.389        & 0.392         & 0.409        & {\ul0.330}          & {\ul0.371}   & 0.342           & 0.379           & 0.354            & 0.384     \\
\multicolumn{1}{c|}{}                          & 96  & \re{\textbf{0.364}} & \re{{\ul 0.392}} & {\ul 0.373}   & \textbf{0.391}  & 0.398         & 0.440        & 0.427         & 0.430        & 0.374          & 0.394   & 0.386           & 0.405           & 0.395            & 0.407     \\
\multicolumn{1}{c|}{}                          & 192 & \re{\textbf{0.421}} & \re{{\ul0.422}} & {\ul 0.427}   &  \textbf{0.421} & 0.451         & 0.440        & 0.465         & 0.452        & 0.429          & 0.429   & 0.441           & 0.436           & 0.445            & 0.434     \\ \cmidrule{2-16} 
\multicolumn{1}{c|}{}                          & Avg & \re{\textbf{0.343}} & \re{\textbf{0.378}} & 0.351   &  {\ul 0.380}  & 0.376         & 0.398        & 0.405         & 0.416        & {\ul 0.349}          & 0.382   & 0.360           & 0.390           & 0.372            & 0.393     \\ \midrule
\multicolumn{1}{c|}{\multirow{6}{*}{ETTh2}}    & 24  & \re{\textbf{0.167}} & \re{\textbf{0.259}} & {\ul 0.171}   & {\ul 0.262}  & 0.182         & 0.272        & 0.178         & 0.269        & 0.172          & 0.263   & 0.186           & 0.276           & 0.173            & 0.265     \\
\multicolumn{1}{c|}{}                          & 36  & \re{\textbf{0.200}} & \re{\textbf{0.272}} & {\ul 0.203}   & {\ul 0.281}  & 0.211         & 0.292        & 0.210         & 0.290        & 0.206          & 0.287   & 0.217           & 0.297           & 0.204            & 0.285     \\
\multicolumn{1}{c|}{}                          & 48  & \re{\textbf{0.220}} & \re{\textbf{0.294}} & {\ul 0.225}   & {\ul 0.298}  & 0.232         & 0.302        & 0.235         & 0.306        & 0.230          & 0.305   & 0.242           & 0.313           & 0.226            & 0.300     \\
\multicolumn{1}{c|}{}                          & 96  & \re{\textbf{0.278}} & \re{\textbf{0.332}} & {\ul 0.286}   & {\ul 0.336}  & 0.295         & 0.346        & 0.300         & 0.348        & 0.301          & 0.352   & 0.297           & 0.349           & 0.291            & 0.341     \\
\multicolumn{1}{c|}{}                          & 192 & \re{\textbf{0.356}} & \re{\textbf{0.381}} & {\ul 0.363}   & {\ul 0.387}  & 0.386         & 0.399        & 0.374         & 0.398        & 0.386          & 0.406   & 0.380           & 0.400           & 0.377            & 0.393     \\ \cmidrule{2-16} 
\multicolumn{1}{c|}{}                          & Avg & \textbf{0.245} & \re{\textbf{0.308}} & {\ul 0.250}   & {\ul 0.313}  & 0.261         & 0.322        & 0.259         & 0.322        & 0.259          & 0.323   & 0.264           & 0.327           & 0.254            & 0.317     \\ \midrule
\multicolumn{1}{c|}{\multirow{6}{*}{Weather}}  & 24  & \re{\textbf{0.101}} & \re{\textbf{0.131}} & {\ul 0.105}   & {\ul 0.136}  & 0.106         & 0.140        & 0.107         & 0.142        & 0.110           & 0.143   & 0.107           & 0.140            & 0.105            & 0.137     \\
\multicolumn{1}{c|}{}                          & 36  & \re{\textbf{0.119}} & \re{\textbf{0.150}} & {\ul 0.122}   & {\ul 0.156}  & 0.124         & 0.161        & 0.126         & 0.166        & 0.134          & 0.172   & 0.125           & 0.159           & 0.126            & 0.163     \\
\multicolumn{1}{c|}{}                          & 48  & \re{\textbf{0.130}} & \re{\textbf{0.166}} & {\ul 0.135}   & {\ul 0.171}  & 0.142         & 0.154        & 0.140         & 0.182        & 0.148          & 0.189   & 0.136           & 0.173           & 0.141            & 0.181     \\
\multicolumn{1}{c|}{}                          & 96  & \re{\textbf{0.163}} & \re{\textbf{0.207}} & {\ul 0.167}   & {\ul 0.211}  & 0.195         & 0.233        & 0.174         & 0.220        & 0.183          & 0.223   & 0.174           & 0.214           & 0.177            & 0.218     \\
\multicolumn{1}{c|}{}                          & 192 & \re{\textbf{0.208}} & \re{\textbf{0.247}} & {\ul 0.212}   & {\ul 0.253}  & 0.240         & 0.269        & 0.222         & 0.260        & 0.229          & 0.261   & 0.221           & 0.254           & 0.223            & 0.258     \\ \cmidrule{2-16} 
\multicolumn{1}{c|}{}                          & Avg & \re{\textbf{0.144}} & \re{\textbf{0.180}} & {\ul 0.148}   & {\ul 0.185}  & 0.161         & 0.191        & 0.154         & 0.194        & 0.161          & 0.198   & 0.153           & 0.188           & 0.154            & 0.191     \\ \midrule
\multicolumn{1}{c|}{\multirow{6}{*}{Exchange}} & 24  & \re{\textbf{0.022}} & \re{\textbf{0.104}}          & 0.026   & 0.109  & 0.028         & 0.119        & 0.029         & 0.119        & {\ul 0.025}    & {\ul 0.108}   & 0.026           & 0.110           & {\ul 0.025}      & {\ul 0.108}     \\
\multicolumn{1}{c|}{}                          & 36  & \re{\textbf{0.031}} & \re{\textbf{0.124}} & 0.034   & 0.128  & 0.039         & 0.138        & 0.038         & 0.137        & {\ul0.033}          & {\ul0.127}   & 0.036           & 0.133           & 0.034      & 0.129     \\
\multicolumn{1}{c|}{}                          & 48  & \re{\textbf{0.041}} & \re{\textbf{0.142}} & {\ul 0.044}         & {\ul 0.147}  & 0.050         & 0.158        & 0.048         & 0.154        & 0.045          & 0.148   & 0.045           & 0.148           & 0.045      & 0.148    \\
\multicolumn{1}{c|}{}                          & 96  & \re{\textbf{0.082}} & \re{\textbf{0.202}} & 0.086         & 0.208        & 0.090         & 0.210        & 0.091         & 0.211        & {\ul 0.084}          & {\ul
0.206}   & 0.086           & 0.206     & 0.085      & 0.207     \\
\multicolumn{1}{c|}{}                          & 192 & \re{\textbf{0.173}} & \re{\textbf{0.296}} & {\ul 0.177}   & {\ul 0.299}  & 0.181         & 0.302        & 0.182         & 0.304        & 0.179          & 0.300   & {\ul 0.177}     & 0.300           & 0.181            & 0.301     \\ \cmidrule{2-16} 
\multicolumn{1}{c|}{}                          & Avg & \re{\textbf{0.070}} & \re{\textbf{0.174}} & {\ul 0.073}   & 0.178  & 0.078         & 0.185        & 0.078         & 0.185        & {\ul0.073}          & {\ul0.177}   & 0.074           & 0.179           & 0.074            & 0.179     \\ \bottomrule
\end{tabular}}
\label{tab:main}
\end{table*}

\begin{table*}[t]
\centering
\caption{\re{Short-term forecasting performance comparison with input length of 96 and forecasting horizon of 12.}}
\resizebox{0.95\textwidth}{!}{
\begin{tabular}{c|cc|cc|cc|cc|cc|cc|cc}
\toprule
\multirow{2}{*}{Datasets}  & \multicolumn{2}{c|}{TimeKD}     & \multicolumn{2}{c|}{TimeCMA} & \multicolumn{2}{c|}{TimeLLM} & \multicolumn{2}{c|}{UniTime} & \multicolumn{2}{c|}{OFA} & \multicolumn{2}{c|}{iTransformer} & \multicolumn{2}{c}{PatchTST} \\ \cmidrule{2-15}
 & MSE            & MAE            & MSE           & MAE          & MSE           & MAE          & MSE           & MAE          & MSE         & MAE        & MSE             & MAE             & MSE           & MAE          \\ \midrule
PEMS04 & \textbf{0.066} & \textbf{0.165} & {\ul 0.074}   & {\ul 0.178}  & 0.083         & 0.187        & 0.087         & 0.193        & 0.089       & 0.195      & 0.078           & 0.183           & 0.150         & 0.224        \\
PEMS08 & \textbf{0.063} & \textbf{0.161} & {\ul 0.076}   & {\ul 0.180}  & 0.085         & 0.189        & 0.089         & 0.196        & 0.092       & 0.199      & 0.079           & 0.182           & 0.168         & 0.232        \\ \bottomrule
\end{tabular}}
\label{tab:traffic}
\end{table*}

\subsubsection{Baselines.} We compare TimeKD with the following time series forecasting methods that include LLM-based methods, i.e., TimeCMA~\cite{DBLP:journals/corr/abs-2406-01638}, Time-LLM~\cite{jin2023time}, UniTime~\cite{liu2024unitime}, and OFA~\cite{DBLP:conf/nips/ZhouNW0023} and Transformer-based methods, i.e., iTransformer~\cite{DBLP:journals/corr/abs-2310-06625}, PatchTST~\cite{Yuqietal-2023-PatchTST}. 
\begin{itemize}
    \item \textbf{TimeCMA~\cite{DBLP:journals/corr/abs-2406-01638}.} The TimeCMA is an LLM-empowered model that leverages cross-modality alignment to extract effective disentangled features for efficient time series forecasting.
    
    \item \textbf{Time-LLM~\cite{jin2023time}.} The Time-LLM adapts LLMs for time series forecasting by reprogramming the time series with text prototypes. 
    
    \item \textbf{UniTime~\cite{liu2024unitime}.} The UniTime employs LLMs to learn from diverse time series datasets, incorporating pure text instructions for cross-domain time series forecasting.
    
    \item \textbf{OFA~\cite{DBLP:conf/nips/ZhouNW0023}.} The OFA performs time series forecasting by freezing the attention and feed-forward layers in the LLM while fine-tuning other layers.
    
    \item \textbf{iTransformer~\cite{DBLP:journals/corr/abs-2310-06625}.} The iTransformer introduces an inverted embedding to represent long-term dependencies across variables and utilizes the Transformer encoder for time series forecasting.  
    
    \item \textbf{PatchTST~\cite{Yuqietal-2023-PatchTST}.} PatchTST introduces a patching mechanism and a channel-independent strategy with Transformer-based models for time series forecasting.
\end{itemize}

\subsubsection{Evaluation Metrics}
Mean Squared Error (MSE) and Mean Absolute Error (MAE) are adopted as evaluation metrics for forecasting comparison:
\begin{equation}
\text{MSE} = \frac{1}{\mathcal{M}}\sum_{M=1}^{\mathcal{M}} (\mathbf{\hat{X}}_M - \mathbf{X}_M)^2,
\end{equation}
\begin{equation}
    \text{MAE} = \frac{1}{\mathcal{M}}\sum_{M=1}^{\mathcal{M}}|\mathbf{\hat{X}}_M-\mathbf{X}_M|,
\end{equation}
where $\mathcal{M}$ is the number of all predicted values. 

\subsubsection{Implementation Details} To enable fairness, the \textit{test batch size is set to 1} for all methods in the testing phase to avoid the drop last batch trick~\cite{DBLP:journals/corr/abs-2403-20150}. All models follow the same experimental setup with an input length of $96$ and forecasting horizons of $24, 36, 48, 96$, and $192$ across datasets. We train our method using the AdamW optimizer and select the trained model with the lowest average validation loss for testing. We utilize the BERT, GPT-2, and LLaMA-3.2 as large language models. Each experiment is repeated three times with different seeds on NVIDIA A100 GPUs. We set different values to other hyperparameters, e.g., the number of LLM layers, the hidden dimension of the Transformer, and the number of Transformer layers are set to 12, 64, and 2, respectively. Please refer to the associated code repository for details\footnote{https://github.com/ChenxiLiu-HNU/TimeKD}.

\subsection{Experiment Results}
\subsubsection{Long-term Forecasting Performance Comparison}
We report the MSE and MAE values of the methods in Table~\ref{tab:main}. The best performance by an existing method is underlined, and the overall best performance is marked in bold. The following observations are made.
\begin{itemize}
    \item \re{TimeKD achieves the best results on all datasets across all forecasting horizons ($FH \in \{24, 36, 48, 96, 192\}$). TimeKD performs better than the best among the baselines up to 9.11\% and 7.52\% in terms of MSE and MAE, respectively. This is attributed to TimeKD's privileged knowledge distillation, which leverages the calibrated LLMs to extract temporal features while mitigating textual noise in the embeddings. We also observe that the performance improvements on ETTm2 exceed those on other datasets because ETTm2 has a higher sampling frequency and finer-grained numerical records. This verifies TimeKD's ability to learn robust representations for frequently sampled datasets.}
    \item Generally, LLM-based methods, benefiting from the powerful cross-domain knowledge of LLMs, perform better than Transformer-based methods in most cases. It shows the superiority of LLM's generic knowledge and powerful knowledge transfer capabilities. iTransformer performs the worst on all datasets especially on the ETT datasets with fewer variables, as it has a simple model structure without sufficient parameters.
    \item TimeCMA performs the best among the existing methods due to its prompts that abstract better temporal trends and the design of cross-modality alignment for retrieving robust time series representations.
\end{itemize}

\re{\subsubsection{Short-term Forecasting Performance Comparison}
As shown in Table~\ref{tab:traffic}, TimeKD achieves the best performance among all baselines on the PEMS04 and PEMS08 datasets. Specifically, TimeKD outperforms TimeCMA with 10.81\% and 10.26\% reductions in MSE on PEMS04 and PEMS08, respectively. The third-best model, iTransformer, is surpassed by TimeKD with 15.38\% and 11.39\% improvements on PEMS04 and PEMS08, respectively.
The superior results of the top three models (TimeKD, TimeCMA, and iTransformer) can be attributed to their use of inverted embeddings, which effectively capture the spatial dependencies among traffic sensors. In contrast, Time-LLM, UniTime, OFA, and PatchTST handle each sensor independently, without considering spatial interactions, and thus achieve relatively lower performance.}

\subsubsection{Ablation Studies of Model Design}
To gain insight into the effects of the different components of TimeKD, including privileged information, calibrated attention, language model, subtractive cross attention, correlation distillation, and feature distillation. We evaluate the following variants.

\begin{itemize}
    \item \textbf{w/o\_PI}. TimeKD without privileged information (e.g., ground truth prompts), where only feed historical data into the teacher model. \re{This variant is the traditional teacher in Figure~\ref{fig:teachers}.}
    \item \textbf{w/o\_CA}. TimeKD without calibrated attention, using the original multi-head attention mechanism. 
    \item \textbf{w/o\_CLM}. TimeKD without calibrated language models, where the teacher model does not leverage LLMs for textual prompt encoding.
    \item \textbf{w/o\_SCA}. TimeKD without subtractive cross attention, using the direct subtraction of embeddings replaces the subtractive cross attention.
    \item \textbf{w/o\_CD}. TimeKD without correlation distillation,  eliminating direct interactions between the privileged transformer and the time series transformer.
    \item \textbf{w/o\_FD}. TimeKD without the feature distillation, removing alignments between the outputs of the privileged transformer and the time series transformer.
\end{itemize}

\begin{figure}[t]
\centering
    \begin{minipage}[b]{0.49\linewidth}
        \centering
        \includegraphics[width=\linewidth]{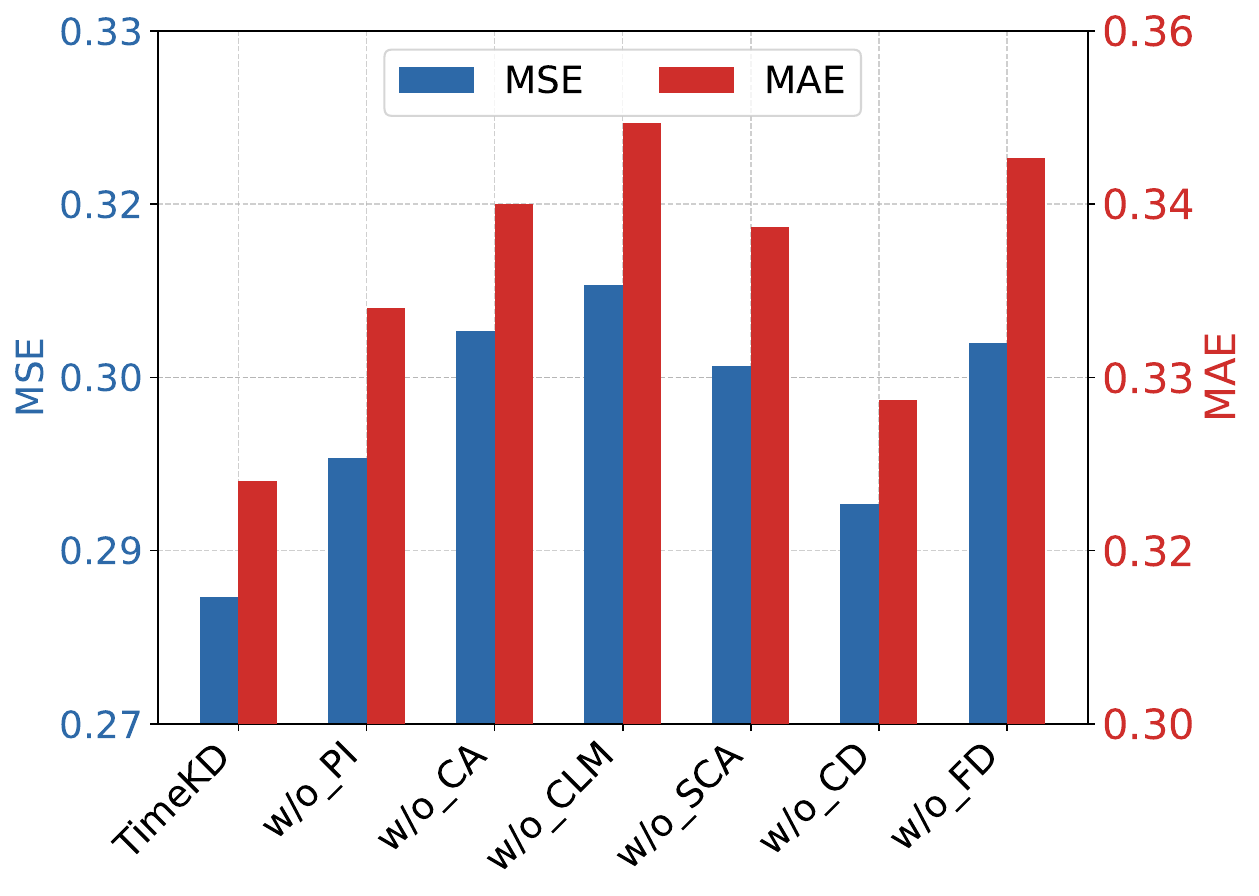}
        \caption*{(a) ETTm1.}
    \end{minipage}
    \hfill
    \begin{minipage}[b]{0.49\linewidth}
        \centering
        \includegraphics[width=\linewidth]{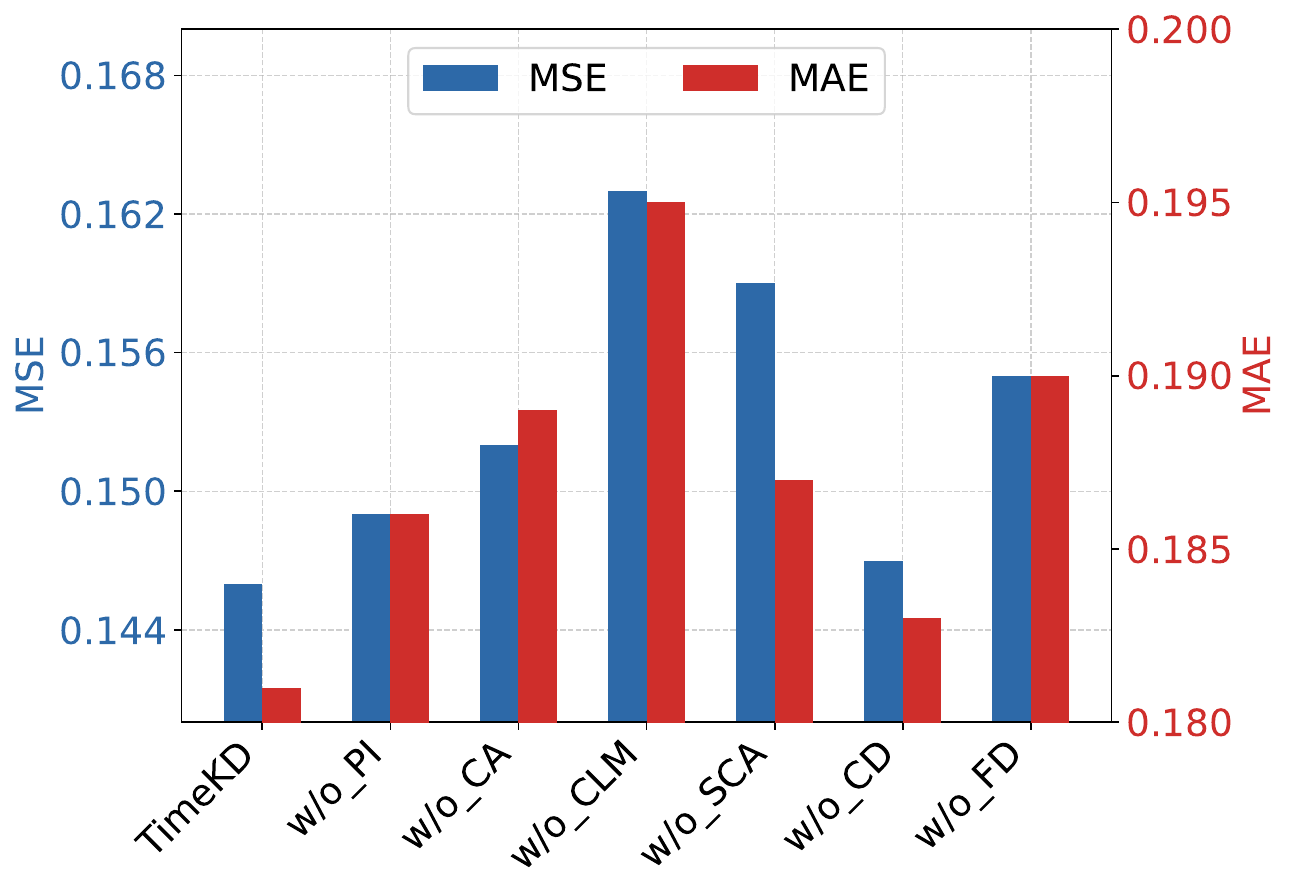}
        \caption*{(b) Weather.}
    \end{minipage}
    \hfill
    \begin{minipage}[b]{0.49\linewidth}
        \centering
        \includegraphics[width=\linewidth]{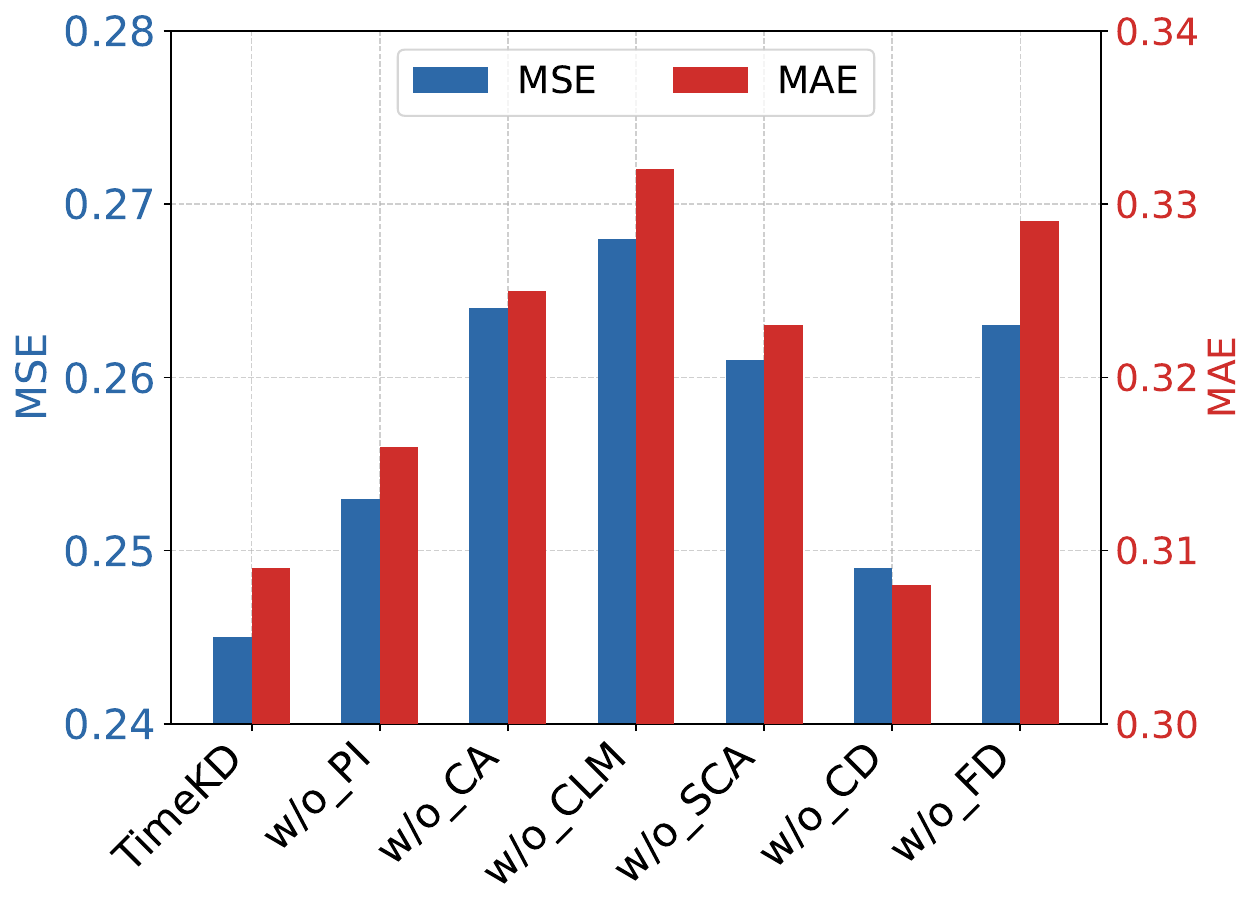}
        \caption*{(c) ETTh2.}
    \end{minipage}
        \hfill
    \begin{minipage}[b]{0.495\linewidth}
        \centering
        \includegraphics[width=4.45cm,height=3.1cm]
        {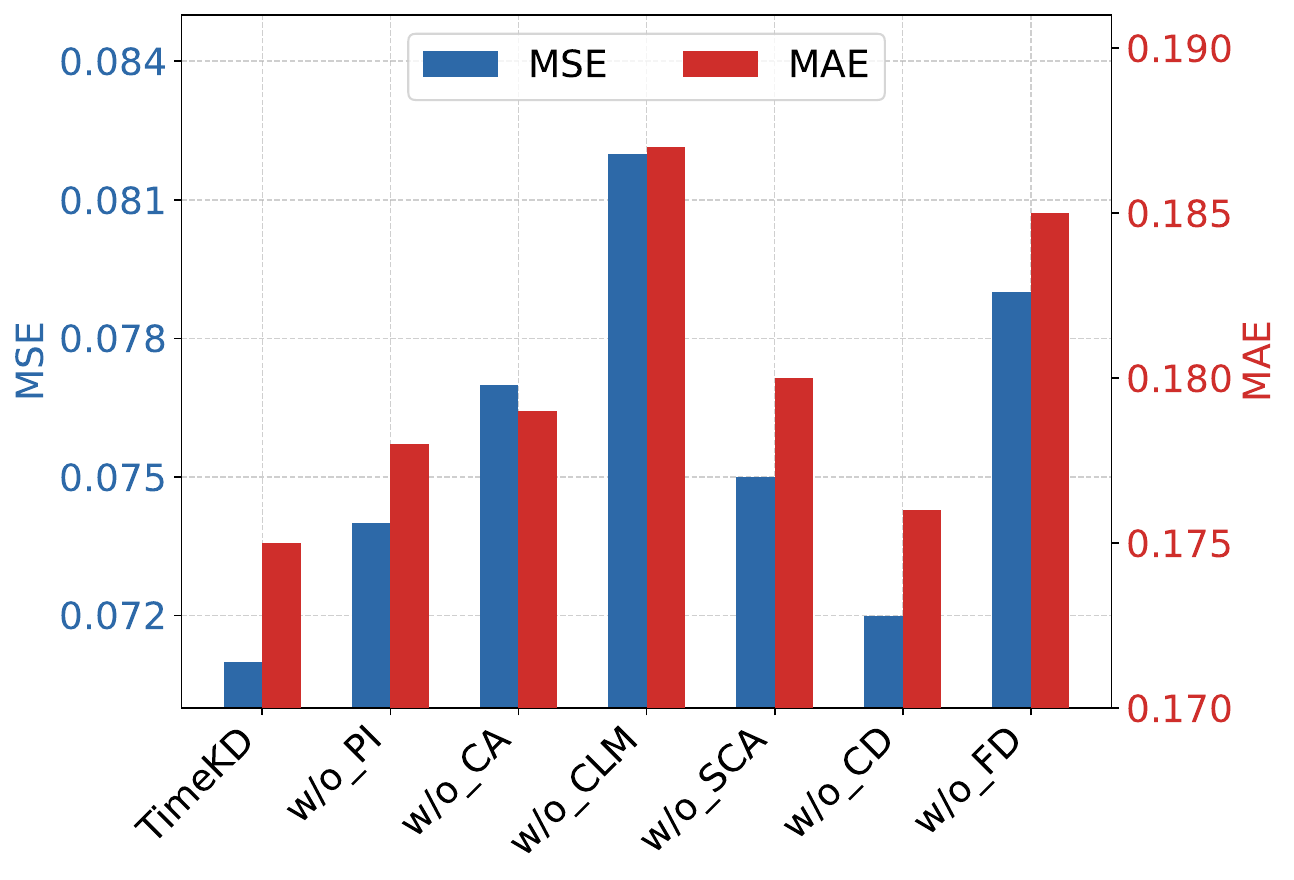}
        \caption*{(d) Exchange.}
    \end{minipage}
    \caption{Performance of TimeKD and its variants on Four Datasets.}
    \label{fig:ablation}
\end{figure}

Figure~\ref{fig:ablation} shows the average results on ETTm1, ETTh2, Weather, and Exchange datasets across all forecasting horizons. Regardless of the datasets, TimeKD outperforms its counterparts without CD and PI. This shows that correlation distillation and privileged information are useful for effective time series forecasting. Notably, when the teacher model only includes historical data without the ground truth prompt, the performance of TimeKD is reduced, which indicates the significance of the privileged information during the distillation. In the ETTm1, TimeKD achieves MSE and MAE reductions by up to 8.2\% and 6.5\%, respectively, compared with w/o\_SCA. In addition, TimeKD performs better than w/o\_CA by at least 8.9\% and 8.4\% in terms of MSE and MAE, respectively, indicating the effectiveness of the subtractive cross attention and calibrated attention. Further, w/o\_FD underperforms compared to most other variants, while w/o\_CLM demonstrates the weakest performance among all, emphasizing the contribution of the calibrated language model.

\re{\subsubsection{Ablation Studies of Open-Source LLMs}
Table~\ref{tab:ab_llm} presents an ablation study on LLM selection in TimeKD, comparing three open-source models. BERT serves as a smaller baseline, GPT-2 has slightly more parameters and better prediction results. LLaMA-3.2 achieves the best performance. This trend suggests that larger LLMs capture more complex linguistic patterns, enhancing distillation. However, LLaMA-3.2 provides marginal accuracy gains at a significantly higher computational cost. With a hidden dimension of 4096 versus GPT-2’s 768, it requires more memory and longer inference times. Given GPT-2’s simplicity and strong performance, we adopt it as our TimeKD backbone for better efficiency.}

\begin{table}[t]
\centering
\caption{\re{Ablation study of LLMs within TimeKD on Exchange. The forecasting horizon is 24.}}
\resizebox{0.4\textwidth}{!}{
\begin{tabular}{lc|cc}
\toprule
\multicolumn{2}{c|}{LLMs}                        & \multicolumn{2}{c}{Exchange}    \\ \midrule
\multicolumn{1}{l|}{Models}    & Model sizes (B) & MSE            & MAE            \\ \midrule
\multicolumn{1}{l|}{BERT}      & 0.110           & 0.032          & 0.125          \\
\multicolumn{1}{l|}{GPT-2}     & 0.117           & {\ul 0.024}    & {\ul 0.105}    \\
\multicolumn{1}{l|}{LLaMA-3.2} & 3               & \textbf{0.020} & \textbf{0.102} \\ \bottomrule
\end{tabular}}
\label{tab:ab_llm}
\end{table}

\subsubsection{Resource Efficiency} 

\begin{table}[t]
\centering
\caption{Efficiency analysis of TimeKD and baselines. The forecasting horizon is 96}
\label{tab:efficiency}
\resizebox{0.5\textwidth}{!}{
\begin{tabular}{c|cccc}
\toprule
\multirow{2}{*}{Models} & \multicolumn{4}{c}{ETTm1} \\ 
\cmidrule(lr){2-5}
 & Trainabl. Param. & \re{Train. Time} & Mem. & Infer. Speed \\ 
\midrule
iTransformer & \textbf{0.22}  & \re{\textbf{12.97}}  & 1,722  & \underline{0.08} \\
Time-LLM     & 44.66          & \re{4799.64}         & 28,882 & 1.08 \\
UniTime      & 108.54         & \re{2472.15}         & 4,168  & 0.39 \\
OFA          & 1.75           & \re{425.12}          & 910    & 0.18 \\
TimeCMA      & 17.99         & \re{50.13}          & \underline{821} & 0.09 \\
TimeKD       & \underline{1.72} & \re{\underline{49.78}} & \textbf{730} & \textbf{0.06} \\ 
\bottomrule
\end{tabular}}
\end{table}

\re{We assess the resource efficiency of TimeKD and the baselines on the ETTm1 across four key metrics: Trainable Parameters (Trainabl. Param.) in millions, Training Time (Train. Time) of an epoch in seconds, Memory Usage (Mem.) in MiB, and Inference Speed (Infer. Speed) in seconds per iteration, all evaluated on NVIDIA A100 GPUs. To ensure a fair comparison of memory usage, we set the training batch size to 8 for all baselines, ensuring that each iteration processes exactly 8 samples.}

\re{The efficiency comparison results are presented in Table~\ref{tab:efficiency}:
\textit{(1) TimeKD achieves the lowest memory consumption and the highest inference speed among all methods}, including smaller models like iTransformer. This is due to the design of the last token extractor in the teacher model and the simplified hyperparameter settings in the student model, which has already benefited from learning pre-trained knowledge from the LLM.
\textit{(2) TimeKD has the lowest trainable parameter count and training time among the LLM-based methods}, and is the second only to iTransformer. This is due to the additional learnable parameters introduced by the privileged Transformer in the teacher model and the privileged knowledge distillation, which result in slightly higher training costs compared to the standalone student model.}

\subsubsection{Scalability}

\begin{figure}[t]
\centering
    \begin{minipage}[b]{0.49\linewidth}
        \centering
        \includegraphics[width=\linewidth]{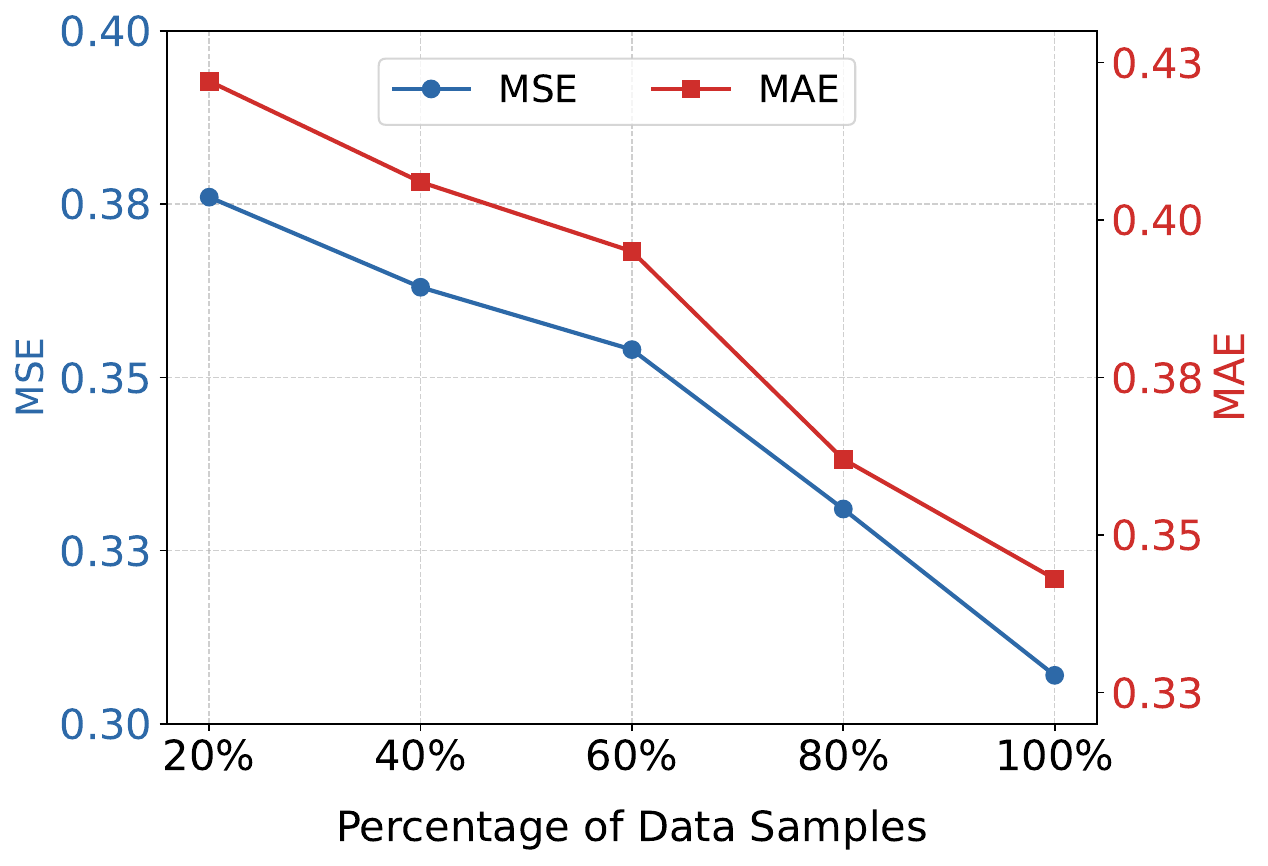}
        \caption*{(a) ETTm1.}
    \end{minipage}
    \hfill
    \begin{minipage}[b]{0.49\linewidth}
        \centering
        \includegraphics[width=\linewidth]{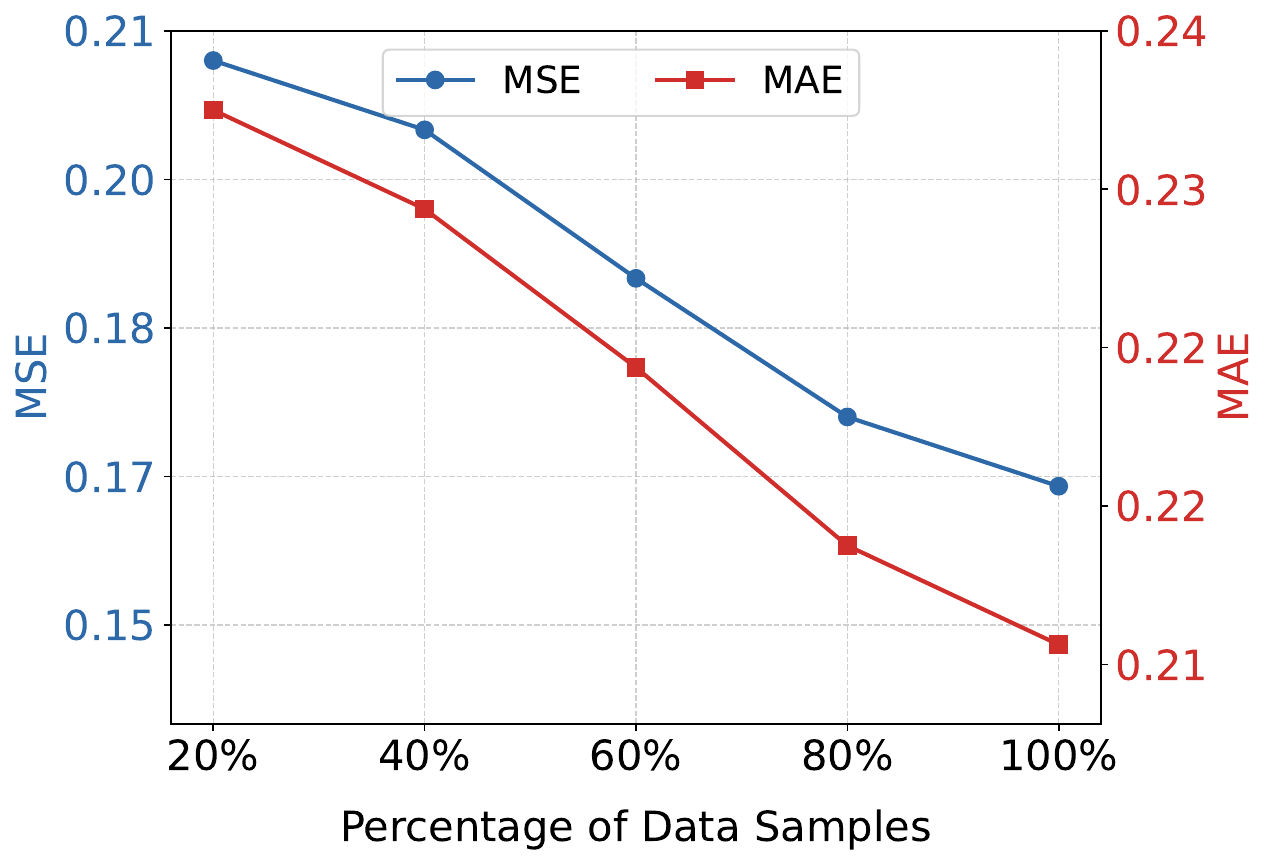}
        \caption*{(b) Weather.} 
    \end{minipage}
    \hfill
    \begin{minipage}[b]{0.49\linewidth}
        \centering
        \includegraphics[width=\linewidth]{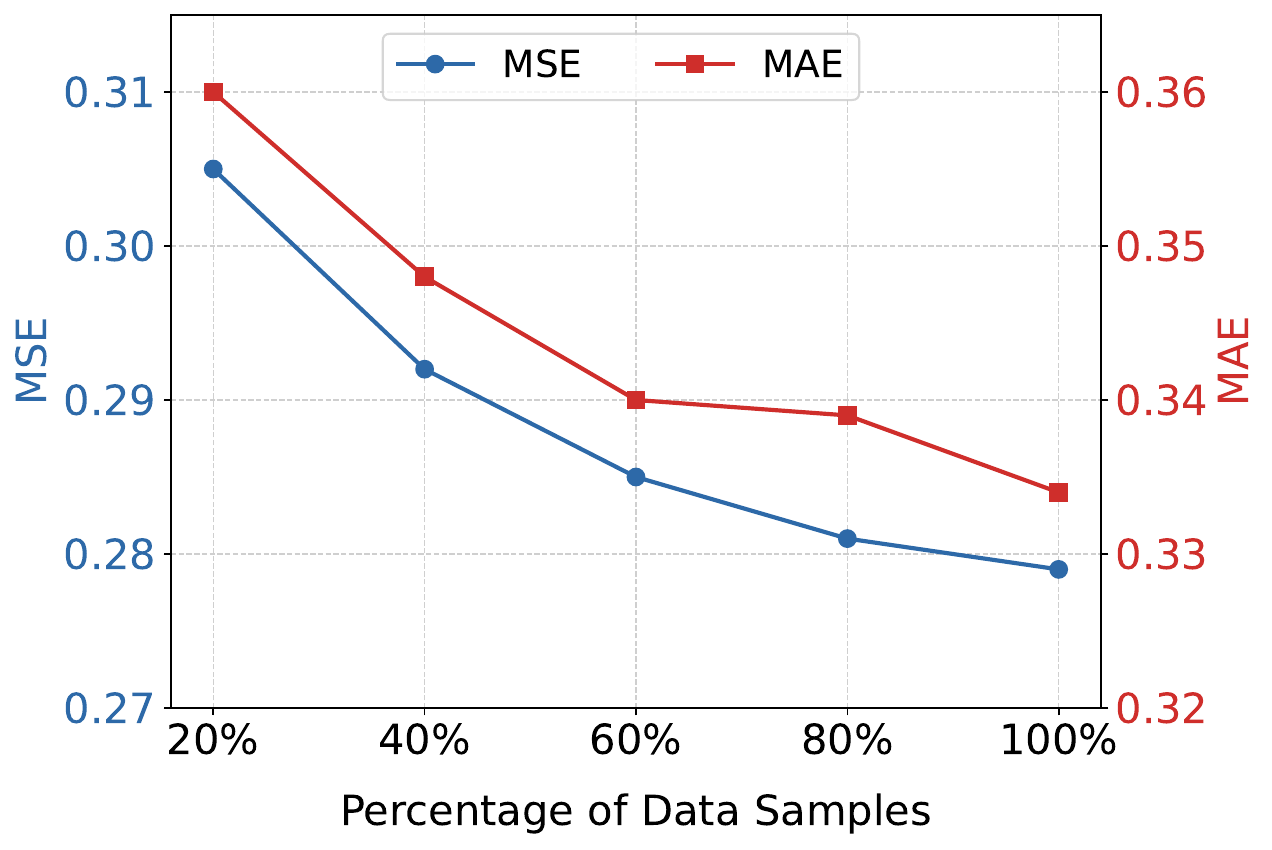}
        \caption*{(c) ETTh2.} 
    \end{minipage}
        \hfill
    \begin{minipage}[b]{0.49\linewidth}
        \centering
        \includegraphics[width=\linewidth]{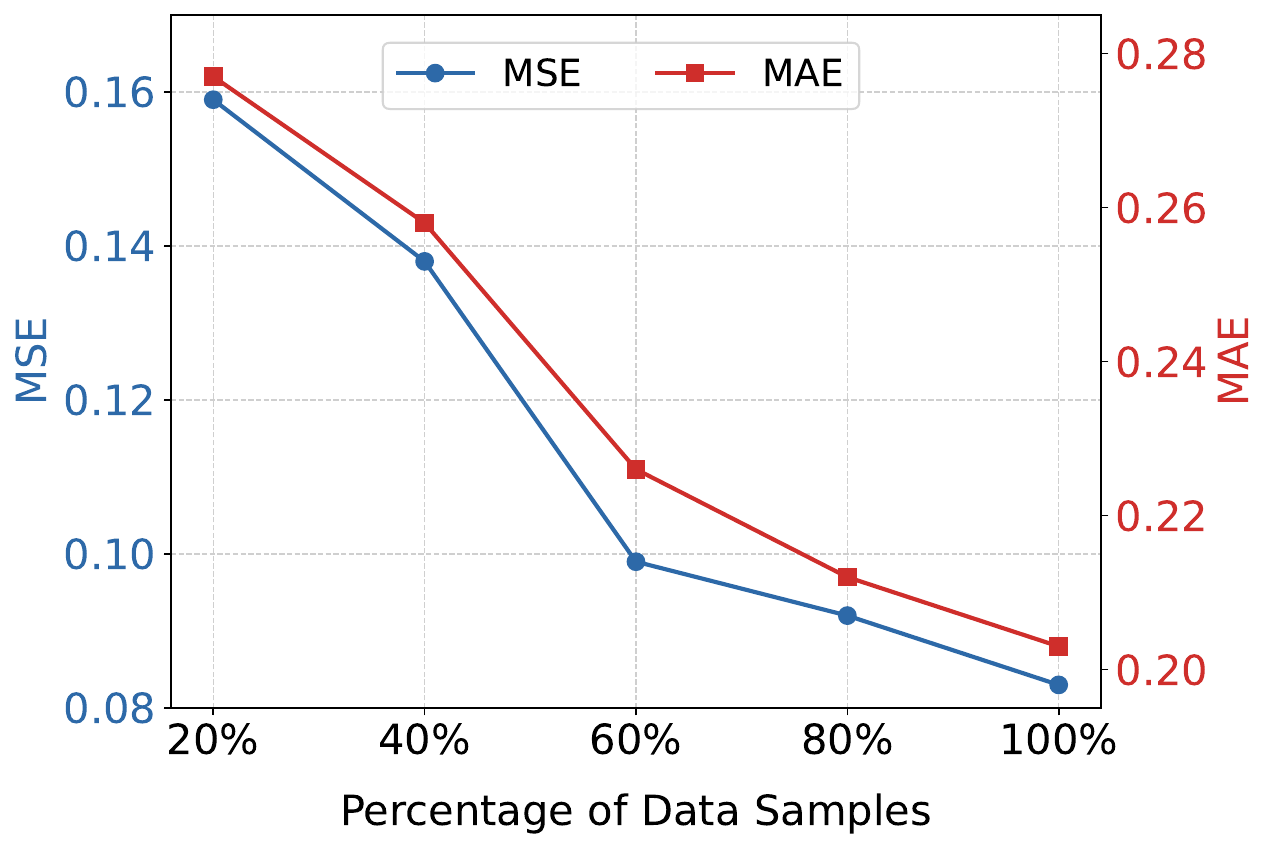}
        \caption*{(d) Exchange.} 
    \end{minipage}
    \caption{Effect of Different Training Data on Four Datasets.}
    \label{fig:scalability}
\end{figure}

The scalability analysis of TimeKD on ETTm1, ETTh2, Weather, and Exchange datasets with a forecasting horizon set to 96 demonstrates its robustness under different data scarcity scenarios, as shown in Figure \ref{fig:scalability}. As the percentage of available data samples increases from 20\% to 100\%, MSE and MAE consistently decrease for both datasets. This indicates that TimeKD effectively leverages the additional data to improve forecasting accuracy, showing adaptability to varying data scales. Notably, the performance improvement is more pronounced in the ETTm1, reflecting its ability to capture complex temporal patterns in frequently sampled data. Similarly, for the Weather, the steady decline in the metrics highlights the capability of TimeKD to generalize across diverse datasets. These results confirm the scalability of TimeKD under limited data conditions, making it suitable for real-world applications with varying data availability.

\subsubsection{Few-shot forecasting}

\begin{table*}[t]
\centering
\caption{Few-shot forecasting on 10\% training data. The input length and forecasting horizon are set to 96 time steps.}
\setlength{\tabcolsep}{2.8mm}{
\begin{tabular}{c|cc|cc|cc|cc|cc|cc|cc}
\toprule
\multirow{2}{*}{Datasets}  & \multicolumn{2}{c|}{TimeKD}     & \multicolumn{2}{c|}{TimeCMA} & \multicolumn{2}{c|}{Time-LLM} & \multicolumn{2}{c|}{UniTime} & \multicolumn{2}{c|}{OFA} & \multicolumn{2}{c|}{iTransformer} & \multicolumn{2}{c}{PatchTST} \\ \cmidrule{2-15}
 & MSE            & MAE            & MSE           & MAE          & MSE           & MAE           & MSE           & MAE          & MSE         & MAE        & MSE             & MAE             & MSE           & MAE          \\ \midrule
ETTm1  & \textbf{0.429} & \textbf{0.396} & {\ul 0.442}   & {\ul 0.439}  & 0.587         & 0.491         & 0.559         & 0.486        & 0.615       & 0.497      & 0.565           & 0.484           & 0.558         & 0.478        \\
ETTm2  & \textbf{0.183} & \textbf{0.261} & {\ul 0.185}   & {\ul 0.265}  & 0.189         & 0.270         & 0.186         & 0.267        & 0.187       & 0.266      & 0.194           & 0.277           & 0.189         & 0.268        \\
ETTh1  & \textbf{0.421} & \textbf{0.415} & {\ul 0.431}   & {\ul 0.422}  & 0.498         & 0.462         & 0.502         & 0.467        & 0.462       & 0.449      & 0.537           & 0.493           & 0.433         & 0.428        \\
ETTh2  & \textbf{0.304} & \textbf{0.341} & {\ul 0.314}   & 0.355        & 0.329         & 0.367         & 0.331         & 0.368        & 0.327       & 0.359      & 0.341           & 0.378           & {\ul 0.314}   & {\ul 0.354}  \\ \bottomrule
\end{tabular}}
\label{table:fewshot}
\end{table*}

To demonstrate whether the proposed TimeKD has remarkable few-shot learning capabilities, we evaluate TimeKD and existing methods with limited training data, i.e., the first $10\%$ of the training data, and set the forecasting horizon to $96$. The prediction results are given in Table~\ref{table:fewshot}. Overall, TimeKD consistently surpasses other competitive baselines, indicating its stable and superior performance, especially on ETTm1. Specifically, TimeKD performs better than TimeCMA by up to {3.18\%} and {9.79\%} in terms of MSE and MAE, respectively, illustrating the importance of effective knowledge distillation in enhancing model performance under data scarcity. The LLM-based method TimeCMA achieves the best performance among existing methods (except MAE on the ETTh2). Furthermore, LLM-based methods perform better than Transformer-based methods in most scenarios. The reason is that these LLM-based methods leverage the pre-trained knowledge of large language models, enabling them to capture temporal patterns even with limited training data.  

\subsubsection{Zero-shot forecasting}

\begin{table*}[htbp]
\centering
\caption{Zero-shot Forecasting Results on ETT. The notation ``trained dataset" $\rightarrow$ ``test dataset" indicates the dataset used for training and the one used for testing, respectively. The forecasting horizon is set to 96 time steps.}
\resizebox{\textwidth}{!}{
\begin{tabular}{l|cc|cc|cc|cc|cc|cc|cc}
\toprule
\multicolumn{1}{c|}{Model}  & \multicolumn{2}{c|}{TimeKD}     & \multicolumn{2}{c|}{TimeCMA} & \multicolumn{2}{c|}{Time-LLM} & \multicolumn{2}{c|}{UniTime} & \multicolumn{2}{c|}{OFA} & \multicolumn{2}{c|}{iTransformer} & \multicolumn{2}{c}{PatchTST} \\ \midrule
\multicolumn{1}{c|}{Metric} & MSE            & MAE            & MSE           & MAE          & MSE           & MAE           & MSE           & MAE          & MSE         & MAE        & MSE             & MAE             & MSE           & MAE          \\ \midrule
ETTm1 $\rightarrow$ ETTm2   & \textbf{0.189} & \textbf{0.263} & {\ul 0.192}   & {\ul 0.266}  & 0.194         & 0.267         & 0.197         & 0.273        & 0.196       & 0.267      & 0.202           & 0.279           & 0.193         & 0.269        \\
ETTm2 $\rightarrow$ ETTm1   & \textbf{0.487} & \textbf{0.439} & {\ul 0.495}   & {\ul 0.443}  & 0.514         & 0.441         & 0.529         & 0.449        & 0.519       & 0.457      & 0.679           & 0.52            & 0.554         & 0.460         \\
ETTh1 $\rightarrow$ ETTh2   & \textbf{0.268} & \textbf{0.336} & {\ul 0.295}   & {\ul 0.341}  & 0.300         & 0.346         & 0.308         & 0.349        & 0.297       & 0.344      & 0.296           & 0.344           & 0.296         & 0.343        \\
ETTh2 $\rightarrow$ ETTh1   & \textbf{0.375} & \textbf{0.391} & {\ul 0.389}   & {\ul 0.440}  & 0.522         & 0.479         & 0.537         & 0.490        & 0.512       & 0.472      & 0.575           & 0.515           & 0.485         & 0.461        \\ \bottomrule
\end{tabular}}
\label{tab:zero_shot}
\end{table*}

Beyond few-shot forecasting, LLMs hold potential as effective zero-shot reasoners. We assess the zero-shot capabilities of TimeKD with existing methods. The MSE and MAE values are given in Table~\ref{tab:zero_shot}. We observe that TimeKD outperforms existing methods consistently, showing its powerful zero-shot learning capabilities. Specifically, TimeKD performs better than TimeCMA by up to {9.15\%} and {11.4\%} in terms of MSE and MAE, respectively. The reason is that TimeKD effectively transfers knowledge across datasets through privileged knowledge distillation, enabling it to generalize better in zero-shot scenarios. By extracting common temporal patterns between different datasets, TimeKD enhances its ability to forecast accurately without training directly on the target dataset.

\subsubsection{Attention Maps Visualization}
\begin{figure}[t]
\centering
    \begin{minipage}[b]{0.49\linewidth}
        \centering
        \includegraphics[width=\linewidth]{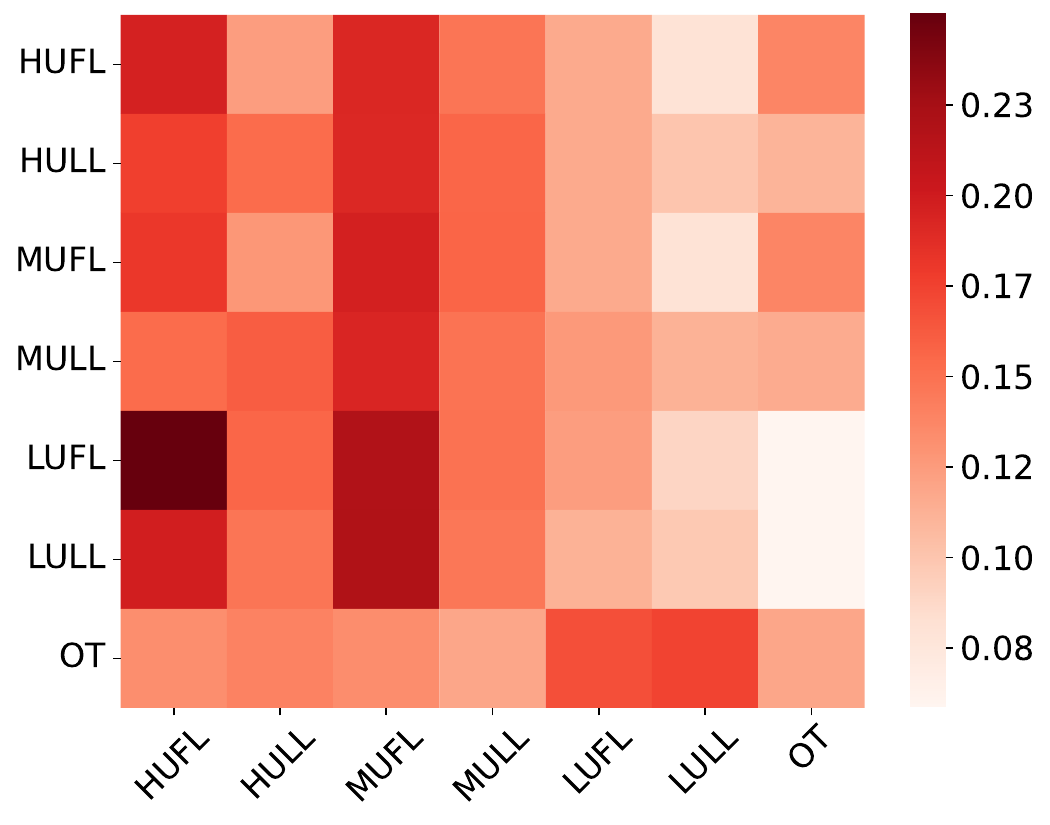}
        \caption*{(a) Privileged Transformer.}
    \end{minipage}
    \hfill
    \begin{minipage}[b]{0.49\linewidth}
        \centering
        \includegraphics[width=\linewidth]{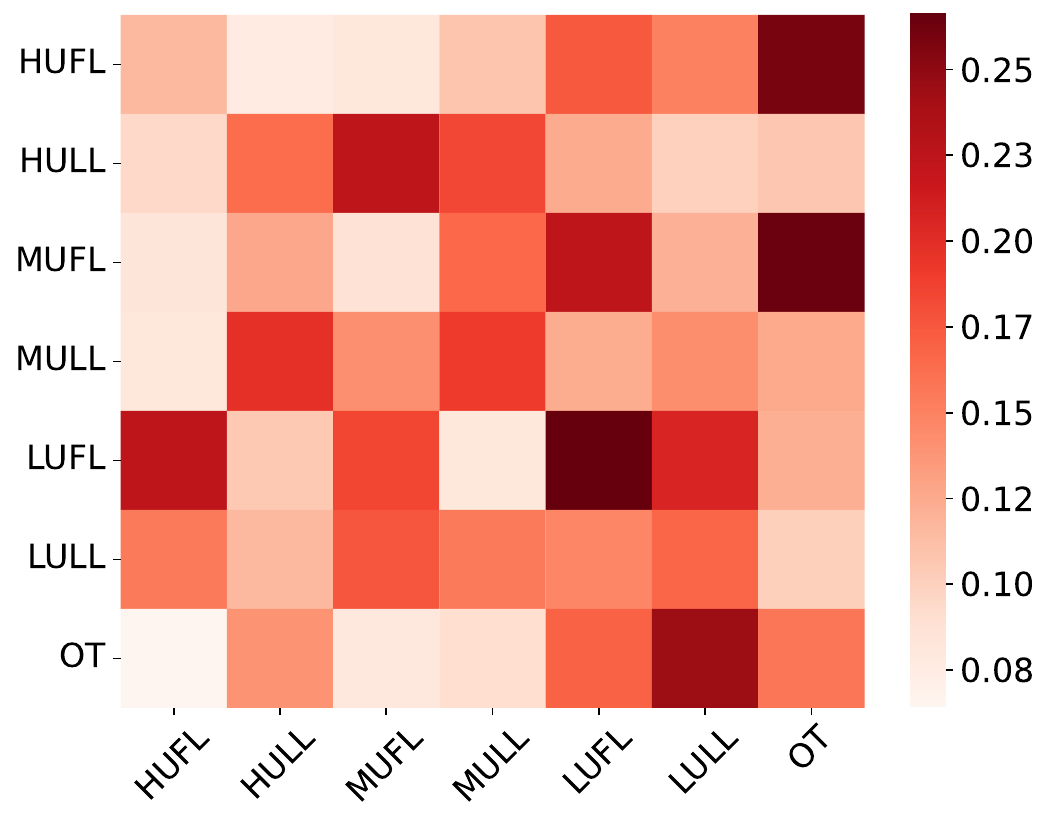}
        \caption*{(b) Time Series Transformer.} 
    \end{minipage}
    \caption{Attention maps from Transformers on ETTm1, capturing the pairwise relations of variables.} 
    \label{fig:trm_at}
\end{figure}
We visualize attention maps from the privileged Transformer and the time series Transformer on the ETTm1 with a forecasting horizon of 96. In Figures~\ref{fig:trm_at} (a) and (b), each row corresponds to a variable and its attention to other variables. The privileged attention is learned from the LLM, while time series attention is extracted from a Pre-LN Transformer encoder. The visualizations highlight distinct behaviors in capturing multivariable dependencies. The LLM textual attention is universal and captures global dependencies between variables due to the pre-trained knowledge from LLMs and privileged information from the input prompts. The attention map from the time series Transformer is local and variable-specific.
By distilling the correlations from the privileged Transformer into the time series Transformer, TimeKD effectively leverages local and global dependencies and enhances forecasting accuracy.

\subsubsection{Feature Visualization}
\begin{figure}[t]
\centering
    \begin{minipage}[b]{0.49\linewidth}
        \centering
        \includegraphics[width=\linewidth]{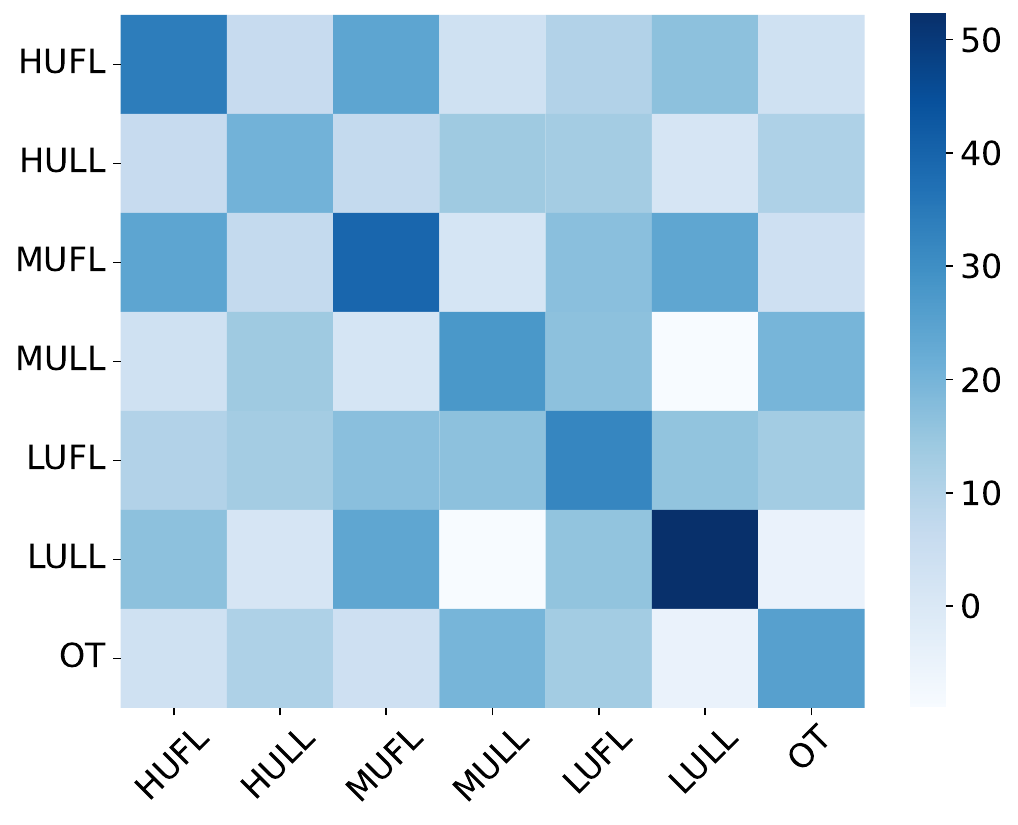}
        \caption*{(a) Privileged Feature.}
    \end{minipage}
    \hfill
    \begin{minipage}[b]{0.49\linewidth}
        \centering
        \includegraphics[width=\linewidth]{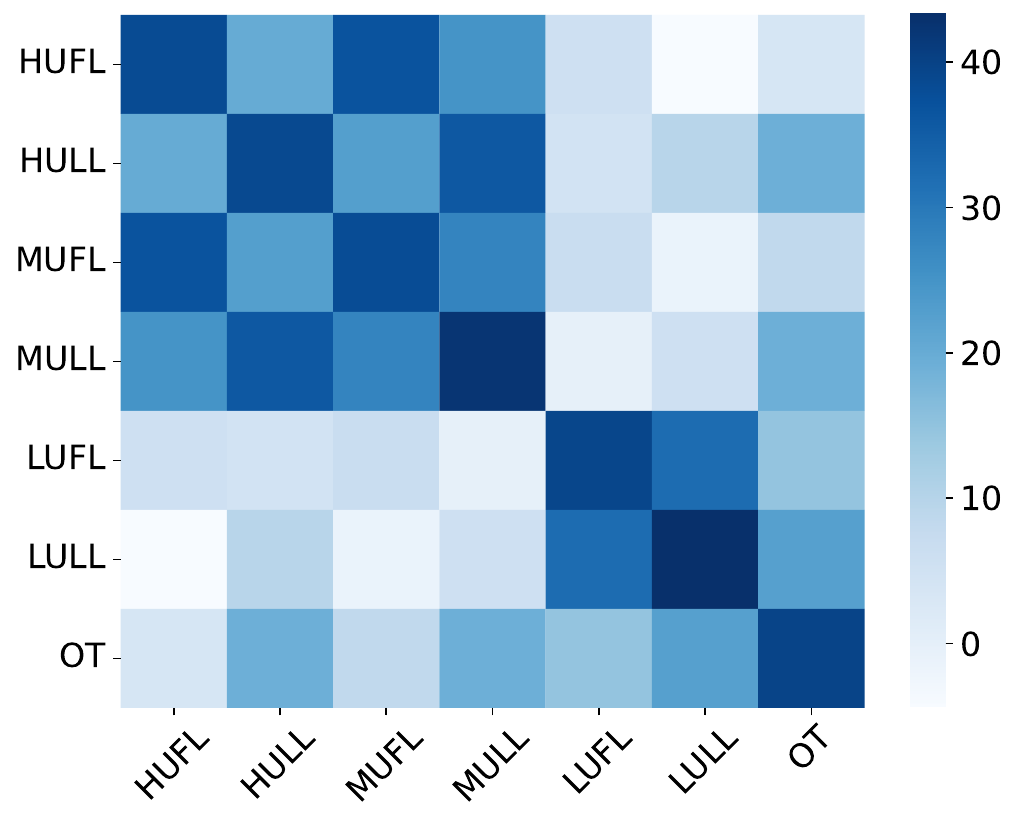}
        \caption*{(b) Time Series Feature.} 
    \end{minipage}
    \caption{Features from the privileged Transformer and the time series Transformer on ETTm1.} 
    \label{fig:trm_feature}
\end{figure}

We visualize feature outputs from the privileged Transformer and time series Transformer on ETTm1 with a forecasting horizon of 96. We multiply the feature matrix by its transpose to compute pairwise interactions between variables, producing a self-relation feature matrix, as shown in Figure \ref{fig:trm_feature}. In Figure \ref{fig:trm_feature} (a), the features are from the privileged Transformer, and show a more comprehensive and balanced pattern of interactions across variables, benefiting from the global contextual knowledge encoded in the LLMs. In contrast, Figure \ref{fig:trm_feature} (b), which visualizes features from the time series Transformer, reveals a localized focus with sparser interactions, reflecting the model's emphasis on local temporal dependencies. 

\subsubsection{Ground Truth vs. Prediction}

\begin{figure}[htbp]
\centering
    \begin{minipage}[b]{0.49\linewidth}
        \centering
        \includegraphics[width=\linewidth]{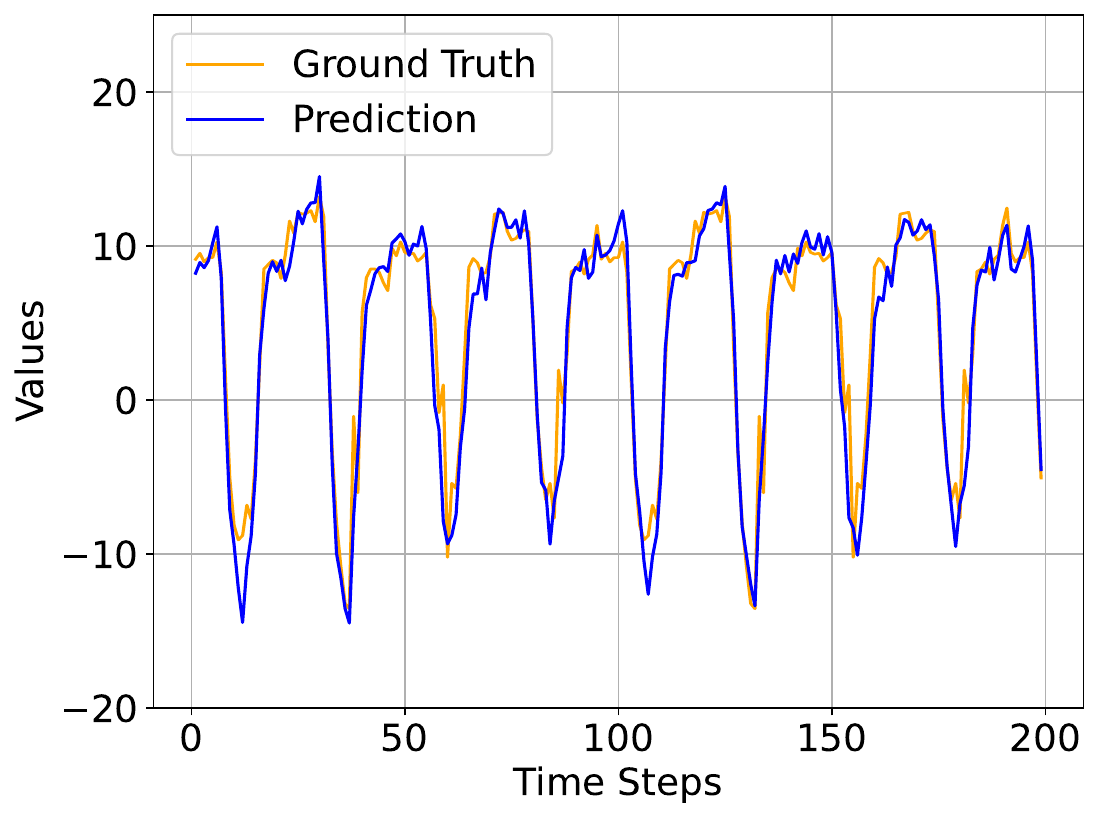}
        \caption*{(a) HUFL.}
    \end{minipage}
    \hfill
    \begin{minipage}[b]{0.49\linewidth}
        \centering
        \includegraphics[width=\linewidth]{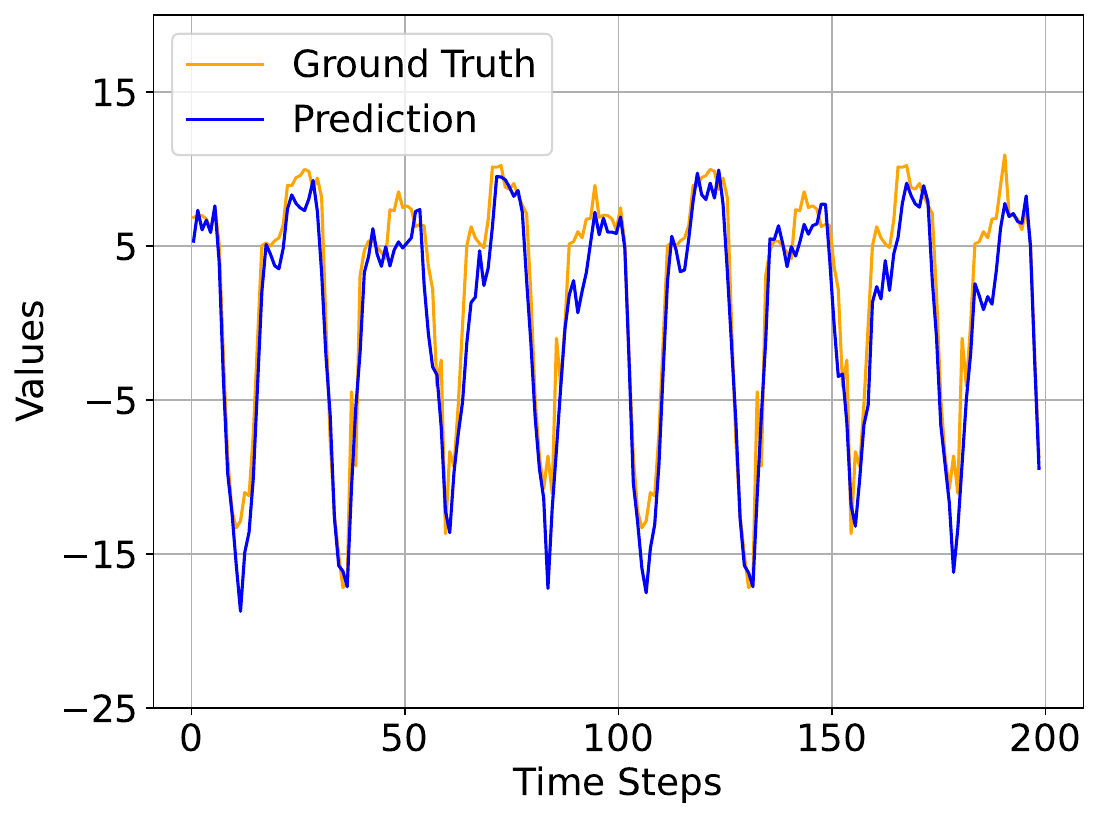}
        \caption*{(b) MUFL.} 
    \end{minipage}
    \hfill
    \begin{minipage}[b]{0.49\linewidth}
        \centering
         \hspace*{0.15cm}
        \includegraphics[width=4.15cm,height=3.39cm]{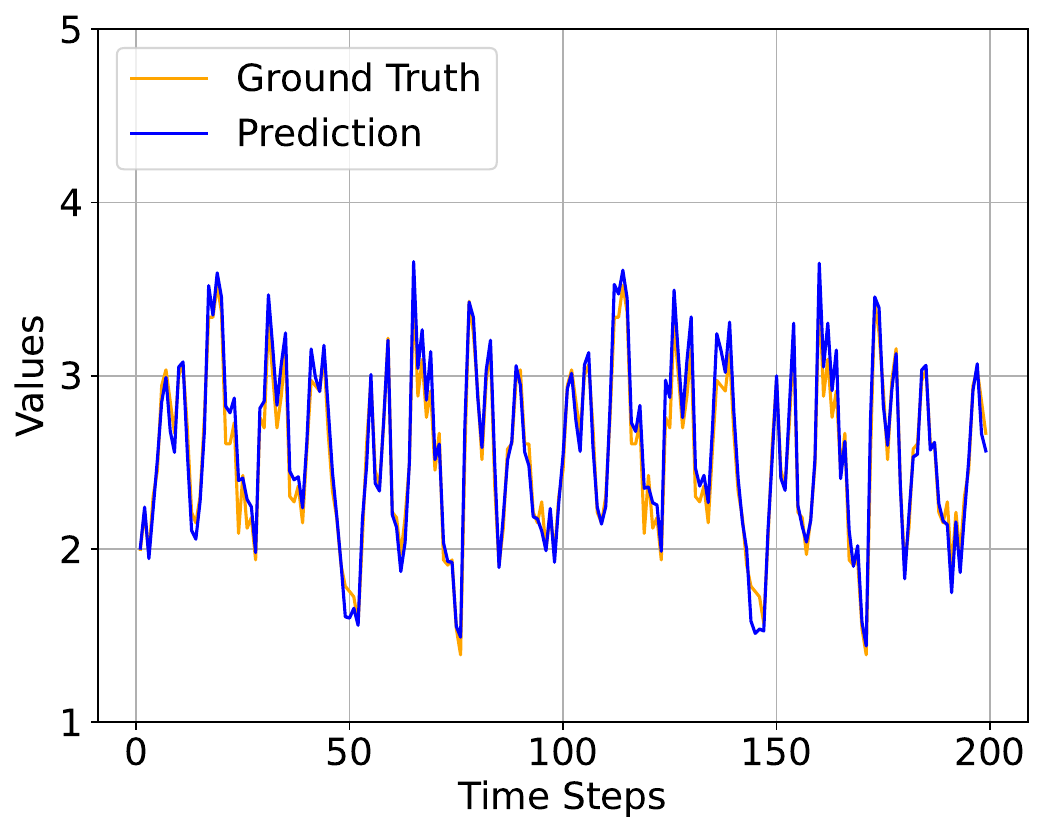}
        \caption*{(c) LUFL.} 
    \end{minipage}
        \hfill
    \begin{minipage}[b]{0.49\linewidth}
        \centering
        \includegraphics[width=\linewidth]{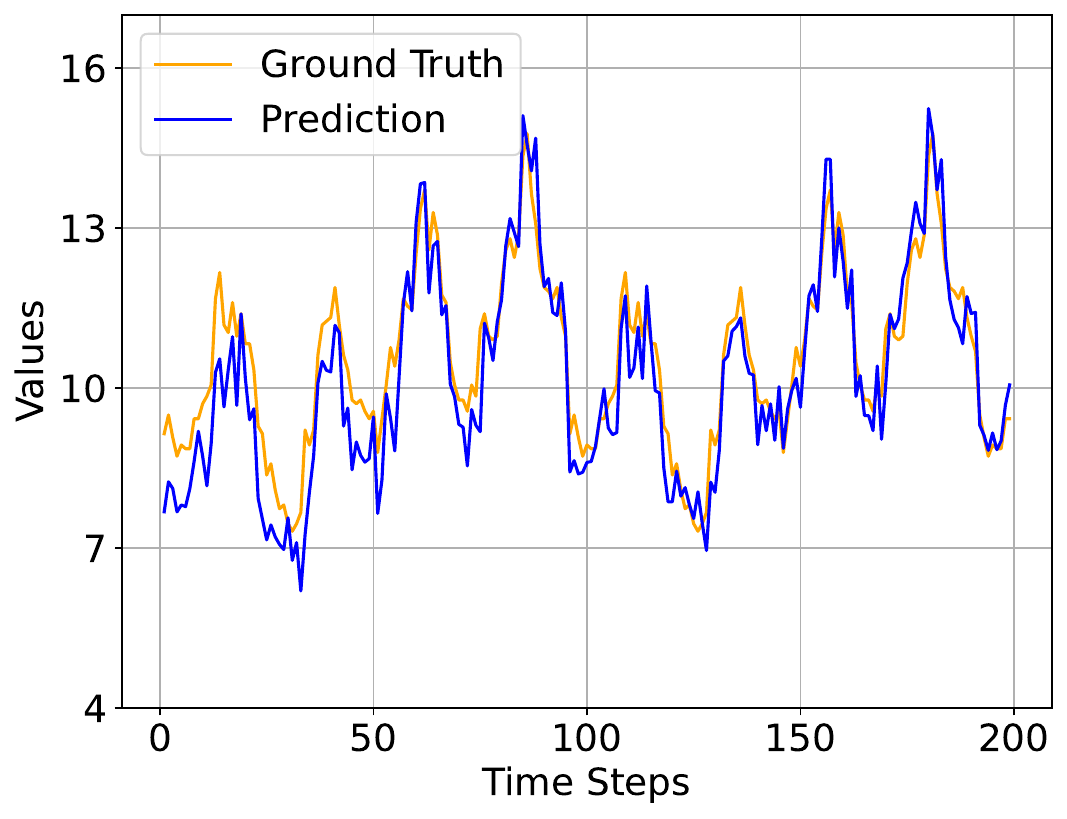}
        \caption*{(d) OT.} 
    \end{minipage}
    \caption{Ground truth vs. prediction visualization on ETTh1.}
    \label{fig:casestudy}
    \vspace{-0.5cm}
\end{figure}

To evaluate whether the TimeKD effectively captures temporal patterns, we visualize the predicted time series and the ground truth on ETTh1, as shown in Figure~\ref{fig:casestudy}. The four subfigures represent the results for the four variables: high useful load (HUFL), middle useful load (MUFL), low useful load (LUFL), and oil temperature (OT), respectively.  From these visualizations, it is evident that the predicted blue curves closely follow the ground truth orange curves, particularly in capturing periodic trends and fluctuations over time. This alignment demonstrates that TimeKD successfully learns and preserves key temporal dependencies, further validating its ability to produce accurate forecasts across multiple time series variables.
\section{Conclusion}
\label{sec:conclusion}

This paper introduces TimeKD, a novel multivariate time series (MTSF) forecasting framework that integrates calibrated language models with privileged knowledge distillation. TimeKD is designed with two key components: a cross-modality teacher model and a lightweight student model. In the cross-modality teacher model, we propose a calibrated language model and subtractive cross-attention. The calibrated language model extracts robust future representations based on the pre-trained knowledge of LLMs and the privileged textual prompts. The subtractive cross attention is proposed to purify the representations to align the time series data. We propose an innovative privileged knowledge distillation including correlation and feature distillation, that transfers the representations from the teacher model to the lightweight student model. We for the first time apply privileged distillation calibration to open-source LLMs for MTSF.
Extensive experiments on real datasets from diverse domains demonstrate the effectiveness and efficiency of TimeKD.

\section{Acknowledgment}
This study is supported under the RIE2020 Industry Alignment Fund – Industry Collaboration Projects (IAF-ICP) Funding Initiative, as well as cash and in-kind contribution from the industry partner(s).
\bibliographystyle{IEEEtran}
\bibliography{References}

\end{document}